\documentclass[twocolumn,10pt]{asme2ej}

\usepackage{amssymb}
\usepackage{graphicx}
\usepackage{caption}
\usepackage{subcaption}
\usepackage{amsmath}
\usepackage{xcolor}



\title{Heterogeneous Metamaterials Design via Multiscale Neural Implicit Representation}

\usepackage{helvet}

\author{Hongrui Chen \qquad Liwei Wang \qquad \textbf{Levent Burak Kara}\thanks{Address all correspondences to lkara@cmu.edu} \affiliation{ Department of Mechanical Engineering\\Carnegie Mellon University\\ Pittsburgh, PA, 15213, USA}}

\begin{document}

\maketitle    

%%%%%%%%%%%%%%%%%%%%%%%%%%%%%%%%%%%%%%%%%%%%%%%%%%%%%%%%%%%%%%%%%%%%%%
\begin{abstract}
{\it 
Metamaterials are engineered materials composed of specially designed unit cells that exhibit extraordinary properties beyond those of natural materials. Complex engineering tasks often require heterogeneous unit cells to accommodate spatially varying property requirements. However, designing heterogeneous metamaterials poses significant challenges due to the enormous design space and strict compatibility requirements between neighboring cells. Traditional concurrent multiscale design methods require solving an expensive optimization problem for each unit cell and often suffer from discontinuities at cell boundaries. On the other hand, data-driven approaches that assemble structures from a fixed library of microstructures are limited by the dataset and require additional post-processing to ensure seamless connections. In this work, we propose a neural network-based metamaterial design framework that learns a continuous two-scale representation of the structure, thereby jointly addressing these challenges. Central to our framework is a multiscale neural representation in which the neural network takes both global (macroscale) and local (microscale) coordinates as inputs, outputting an implicit field that represents multiscale structures with compatible unit cell geometries across the domain, without the need for a predefined dataset. We use a compatibility loss term during training to enforce connectivity between adjacent unit cells. Once trained, the network can produce metamaterial designs at arbitrarily high resolution, hence enabling infinite upsampling for fabrication or simulation. We demonstrate the effectiveness of the proposed approach on mechanical metamaterial design, negative Poisson's ratio, and mechanical cloaking problems with potential applications in robotics, bioengineering, and aerospace.  }
\end{abstract}

\section{INTRODUCTION}
Metamaterials are materials with unit cell architectures engineered to exhibit physical properties not attainable in conventional materials. They have enabled phenomena such as electromagnetic invisibility cloaks, acoustic and thermal wave manipulation, and mechanical behaviors like zero or Negative Poisson’s Ratio (NPR) \cite{fan2021review,yu2018mechanical,jiang2022flexible}. The remarkable properties of metamaterials arise from their designed microstructures rather than the base material composition. This design freedom promises next-generation devices and structures with exceptional, often exotic functionalities. To realize its full potential, concurrent multiscale design is desirable, where each unit cell in a metamaterial is tailored to local requirements while coevolving with the macroscale structures.  However, this multiscale design process is highly challenging due to the vast and complex design space, compounded by the non-trivial task of ensuring compatibility between thousands of unit cells in a large-scale structure. In essence, the flexibility to spatially vary unit cell configurations comes at the cost of a combinatorial design explosion and the need to maintain seamless connectivity between neighboring cells. \footnote{Our code can be found at: https://github.com/HongRayChen/MSNN}

Classical topology optimization approaches for multiscale structures treat the macroscale layout and microscale unit cell design as coupled problems\cite{wu2021topology}. A typical hierarchical formulation will iterate between a macroscale optimization and numerous microscale (unit cell) optimizations for each region of the structure \cite{xia2014concurrent,sivapuram2016simultaneous}. This approach is computationally expensive and prone to compatibility issues, as optimizing each unit cell independently can lead to incompatible adjacent cells, causing mismatched boundaries and discontinuous load paths. Such discontinuities undermine manufacturability and structural performance, requiring additional smoothing or post-processing steps to fix. Due to these challenges, traditional concurrent metamaterial design methods are often time-consuming or forced to restrict design flexibility (for example, using a single repeating cell or a small set of predetermined cell types) to avoid incompatibilities. 

In recent years, data-driven design frameworks have been introduced to alleviate the computational burden of concurrent metamaterial optimization \cite{wang2022mechanical,lee2024data}. In a typical data-driven workflow, a large unit cell library with pre-computed properties is constructed, and the macroscale design optimization then works to distribute those pre-computed properties. Once a target distribution of properties is obtained, corresponding microstructures are retrieved from the database or generated by machine learning to fill each location in the structure. This top-down approach bypasses nested two-scale optimization and can improve design efficiency. However, these methods are constrained by the dataset used and are not easy to go beyond the unit cell geometries or properties available in the library. Ensuring compatibility between adjacent cells remains a critical issue – existing techniques must either enforce similarity between neighboring unit cell shapes or insert specially designed transition regions to seamlessly connect different cells. These extra compatibility measures limit the range of achievable property variations and add complexity to the design process. A clear gap remains for a method that can explore the free-form metamaterial design space while enforcing compatibility across cells in a large-scale structure. 

To address these challenges, we propose a neural network-based approach for concurrent metamaterial design that learns an implicit continuous representation of the two-scale structure. In contrast to post-assembly methods that rely on a discrete library, our approach uses a neural network to directly generate optimized, compatible multiscale structures without the need for post-assembly. Specifically, we extend the concept of coordinate-based neural representations \cite{Chandrasekhar2021,Chandrasekhar2021MM,chen2023cond}(demonstrated in prior work for 2D single-scale designs) to a two-scale setting with a four-dimensional input: two coordinates encode the global position in the macroscale structure, and two coordinates parametrize the local position within a unit cell. The network outputs an implicit density field for any given combination of macro and micro coordinates. In this way, a single neural network compactly represents the entire metamaterial at both scales, implicitly defining an infinite set of unit cells that vary smoothly across the domain. Importantly, neighboring cells have overlapping input coordinates for the neural networks along their boundaries, thus ensuring compatible cells by construction. We further introduce a tailored compatibility loss function during network training to enforce that the material configuration is continuous across cell boundaries. This loss penalizes any geometric discontinuity between adjacent unit cells, effectively making boundary connectivity a learned prior to the design.

\begin{figure*}
\centering

\includegraphics[width=\textwidth]{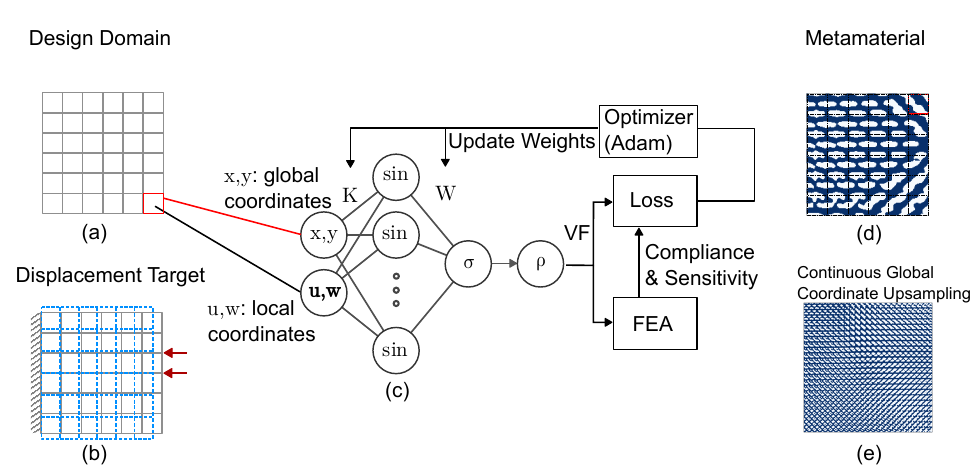}

\caption{ The neural network (c) takes the global $(x,y)$ and local $(u,v)$ coordinates sampled across the design domain (a) as input and outputs density $\rho$ at each coordinate point. The density $\rho$ is then assembled and fed into the finite element analysis and other performance metric analysis to form a combined loss function. We use backpropagation to update the trainable kernel and weights of the neural network. We can obtain the metamaterial (d) by querying the design domain at its original resolution. We can also upsample the global coordinates and solve for a subset of cells during each epoch with mini-epoch solving. This enables us to solve a bigger problem with smoother continuity without a large increase in runtime. (e) Once good continuity is observed, we can then infinitely upsample the global coordinates to obtain an extremely large-scale metamaterial design without compromising the connectivity between cells.  }

\label{fig:flowchart}
\end{figure*}

Our neural network is trained to simultaneously optimize macro-scale performance and meet micro-scale compatibility constraints (Figure \ref{fig:flowchart}). Because the network provides a smooth parametric description of the metamaterial across the unit cells, we no longer need to design each individual unit cell and can instead optimize the weights of the neural network to simultaneously evolve all the unit cells together. It also significantly reduces the number of design variables, as the weights represent a smaller set of parameters compared to assigning a design variable to every microstructural element in a large structure. This reduction in dimensionality enables efficient and scalable solutions to multiscale design problems that would be challenging with traditional methods that use cell-specific variables. Another notable advantage of our approach is the ability to upsample the design to an arbitrarily fine resolution. Since the neural network outputs a continuous geometry, once trained, it can be queried at any level of detail, providing a path to generate high-resolution designs without the need for additional optimization or interpolation. In essence, the trained model acts as a continuous representation of a two-scale structure, meaning a practitioner can zoom in to any region of the structure and obtain a consistent, high-fidelity microstructure geometry. We demonstrate through numerical case studies that this framework can produce smoothly graded metamaterial structures that achieve optimized property distribution for optimal performance. 

The main contributions include: 
\begin{itemize}
    \item A neural-network-based approach for flexible and yet compact representation of the free-form multiscale metamaterial geometry 
    \item A cross-cell blending scheme to enable smooth transitions between neighboring unit cells with distinct geometries
    \item An effective ``mini-epoch'' solving of multiscale design that optimizes a subset of cells every epoch to solve large design domains in reduced runtime 
    \item An efficient multiscale design methodology for heterogeneous metamaterials that enables scalable generation of high-resolution structures
\end{itemize}

\section{RELATED WORK}
In this section, we first introduce representative applications of mechanical metamaterials that will serve as testbeds for our method. We then review existing design approaches relevant to these applications, including concurrent multiscale topology optimization, data-driven multiscale design, and neural-network–parameterized topology optimization. For each approach, we identify key bottlenecks and clarify how our method differs.

\subsection{Applications of Mechanical Metamaterials}
Mechanical metamaterials offer tunability both at the unit-cell level and in how these cells vary across space. Accordingly, their applications can be broadly divided into two categories: designs that focus only on unit-cell architectures, and designs that employ aperiodic assemblies with variations in both unit cells and tiling.

For the former, a typical example is negative Poisson’s ratio (NPR) materials, also known as auxetics. Intuitively, these systems expand laterally when stretched axially (and contract laterally when compressed), providing enhanced indentation resistance, energy absorption, and shear stiffness compared to conventional materials~\cite{evans2000auxetic}. By arranging reentrant or chiral unit cells, NPR metamaterials have been applied in soft robotics, impact protection, and shape-morphing structures~\cite{liu2020two,han2024motion}.

For the latter, the objective is to achieve heterogeneous properties by tailoring aperiodic unit cells to local mechanical requirements. Here, metamaterials act as tunable building blocks distributed according to spatially varying demands such as stiffness or load paths. Representative applications include structural lightweighting and mechanical cloaking. Lightweight designs aim to minimize structural compliance under material usage constraints by allocating appropriate unit cells across regions, a theme long studied via topology optimization and microstructural embedding~\cite{bendsoe1989optimal,wu2021topology}, where smoothly varying layouts can yield high-performing, lightweight structures~\cite{li2018topology}. Cloaking, by contrast, seeks to conceal objects from perturbations in the mechanical field and has been explored across acoustics and elastodynamics to elasticity~\cite{cummer2007one,norris2008acoustic,norris2011elastic}; more recent structural cloaks have also been demonstrated using data-driven multiscale pipelines with separated macro–micro optimization and post-assembly~\cite{wang2022mechanical}.

In this study, we examine all three representative cases—NPR materials, structural lightweighting, and mechanical cloaking—to demonstrate the effectiveness of our method. The NPR case highlights the unit-cell tunability of our approach, the lightweight design illustrates its capability to handle multiscale assemblies, and the cloaking case, being the most complex, demonstrates the generalizability and flexibility of the method.

\subsection{Concurrent Multiscale Topology Optimization}

Concurrent multiscale topology optimization couples macroscale and microscale design decisions in a single framework \cite{wu2021topology,Murphy2021Coupling,Hu2022ThreeScale,Wu2020Hierarchical}, which have been applied to various forms of metamaterials, including free-form, lattice-based, and Triply Periodic Minimal Surfaces (TPMS) \cite{Wang2018Concurrent,Xu2024TPMS}. Rather than optimizing the overall shape first and then filling it with repeated cells, concurrent techniques often treat cell geometry as explicit variables at each finite element or control point, allowing local tailoring of strut‑based lattice systems and graded mesostructures \cite{xia2014concurrent,sivapuram2016simultaneous,garner2019compatibility,Wang2018Concurrent,Liu2023HMMLS,Zhang2023Honeycomb}. The compatibility of adjacent cells can be tackled by enforcing consistent boundary representations across shared interfaces \cite{garner2019compatibility,zhou2008design,du2018connecting,wu2018continuous,alexandersen2015topology,li2018topology,zhou2019level,zhang2018multiscale}, or by extending microstructure boundaries so that continuity is inherently maintained, as in multi‑lattice embeddings that smooth transitions between different lattice systems \cite{Sanders2021Multilattice}. However, the combination of detailed unit-cell design with macroscale spatial variations creates a large nested design space that is computationally expensive to explore. Consequently, some design methods simplify the microscale details using a small set of representative parameters   \cite{wu2019topology,li2018optimal,panesar2018strategies,wang2017concurrent,Xu2024TPMS,lazarov2017maximum,carstensen2018projection,wu2017minimum,schmidt2019structural}. They typically rely on predefined motifs, often specific TPMS or honeycomb families, to guarantee compatibility, albeit with added geometric restrictions. Dehomogenization methods \cite{groen2018homogenization,wu2019design,larsen2018optimal,Clausen2015Hierarchical,Nale2025Embedded} further extend this framework to accommodate continuous and directional tilings of fabricable microstructures, using a postprocessing stage to project elemental densities onto the underlying microstructures. 

Our method builds on concurrent multiscale topology optimization, where macro- and micro-level decisions are coupled in a single problem. Compared with existing approaches, it introduces innovations in both representation and synthesis, yielding three main advantages: (i) parameter efficiency: per-cell design variables are replaced by a shared two-scale network, substantially reducing the number of parameters while preserving expressiveness; (ii) compatibility by construction: neighboring cells share boundary coordinates in the network input and are further regularized by an explicit compatibility loss, thereby eliminating the need for post-hoc smoothing or transition microstructures; and (iii) scalability: a mini-epoch strategy optimizes subsets of cells per epoch while maintaining global coupling, enabling large domains to be optimized within practical runtimes.

\subsection{Data-driven Multiscale Design}
While physics-based topology optimization remains central, data-driven approaches have recently gained momentum as a way to accelerate or refine multiscale design \cite{lee2024data}. In these methods, large databases of candidate microstructures are generated, and their effective properties (e.g., stiffness, thermal conductivity) are pre-computed. Machine learning or surrogate modeling then allows rapid property prediction, bypassing costly full-scale finite element simulations \cite{wu2021topology}. Such data-driven frameworks can facilitate connectivity and compatibility among cells, as the library of microstructures can be pre-filtered or adaptively searched for shapes that mesh well together \cite{liu2020two}. 
Recent advances additionally exploit physics‑augmented neural surrogates for graded lattices \cite{Stollberg2025Graded}, Voronoi‑based surrogate models \cite{Padhy2024Voronoi}, and generative latent spaces for multi‑lattice transitions \cite{Chen2024MulaTOVA}, further broadening the design envelope beyond precomputed catalogs. These surrogate-accelerated approaches improve throughput but introduce approximation error and require task-specific training. Moreover, the achievable performance is bounded by the property envelope of the underlying library. Our method is library-free and self-supervised: the optimal microstructure is optimized in situ during physics-based training of the two-scale implicit field, so the design space is not capped by catalog coverage; interface compatibility is enforced during optimization rather than via similarity heuristics or bespoke transition cells.

\subsection{Direct Topology Optimization with Neural Networks}
We categorize “direct” neural-network-based topology optimization methods as those that do not rely on any precomputed training database. Instead, these methods learn the optimal density distribution in a self-supervised manner. As an early demonstration, Chandrasekhar and Suresh \cite{Chandrasekhar2021} parameterize the density field with a neural network to enable end-to-end topology optimization. The neural network involved in direct topology optimization can be viewed as an implicit neural representation, where some recent works \cite{zhang2021tonr,Zhang2023TOINR,Hu2024IFTONIR,nobari2024nito} have demonstrated promising results on the task of topology optimization. This concept has similarly been extended to the design of metamaterial microstructure cells \cite{sridhara2022generalized}, and we further generalize it to graded multiscale structures by embedding neural representations at each cell across the macroscale design. Recent work also introduced neural implicit representation to level set-based topology optimization \cite{Deng2021PLS, Mallon2025NLS}

To control length-scale requirements, Chandrasekhar and Suresh propose a Fourier-projection-based network \cite{Chandrasekhar2021Fourier}, which has also been applied in multi-material problems \cite{Chandrasekhar2021MM}. Deng and To \cite{deng2020topology} introduce Deep Representation Learning for topology optimization, likewise re-parameterizing the density field with neural networks to address minimum compliance and stress-constrained objectives. In subsequent work \cite{deng2021parametric}, they represent level-set functions with deep neural networks, thereby improving flexibility in describing complex boundaries. Zehnder et al. \cite{zehnder2021ntopo} use multilayer perceptrons to jointly parameterize density and displacement fields in a mesh-free framework, enabling continuous solution spaces learned directly from topology optimization constraints. Mai et al. \cite{mai2023physics} adopt a comparable approach for optimal truss design. A recent work by Chen et al. \cite{chen2024multi} first introduced direct multiscale optimization with neural networks; we further expand this work to explore the application of neural networks as geometry representation to metamaterial design.

\section{PROBLEM FORMULATION}
In this section, we propose the formulation for neural network-based metamaterial design. We begin by giving an overview of the entire pipeline (Figure \ref{fig:flowchart}). In Section 3.1, we first introduce the multiscale network representation. In the following section, we introduce formulations for several metamaterial design cases. We establish the equation for homogenization and inverse homogenization with bulk modulus maximization (Section 3.2). Building upon the equation of homogenization, we introduce the equation for compliance minimization (Section 3.3). Finally, we formulate the equation for metamaterial design with displacement matching (Section 3.4). For displacement matching problems, we add the bulk modulus maximization for each cell to ensure better structural performance. For both types of design problems, we show the combined loss function, which ties all the objectives used for backpropagation to update the parameters of the neural networks.  

\subsection{Neural Network-based Multiscale Representation}
The topology network $T(\textbf{X})$  (Figure \ref{fig:flowchart}) represents the density field as a continuous function of coordinates, in contrast to typical topology optimization methods that represent the density field using discrete pixels or voxels. The topology neural network takes in local domain coordinates $u,w$, as well as the global coordinates $x,y$ for each cell. The global coordinates get concatenated with the local coordinates to form the input to the topology network, $\textbf{X} = [x,y,u,w]$. The local coordinates are normalized between $-0.5$ to $0.5$. The global coordinates are normalized between $-0.5$ to $0.5$ along the longest axis. Given the local and global coordinates, the topology network outputs the density value $\rho$ at each coordinate point. The domain coordinates represent the center of each element in the design domain. During topology optimization, a batch of domain coordinates that correspond to the mesh grid is fed into the topology network. The output is then sent to the Finite Element (FE) solver. The domain coordinates are multiplied by a kernel $\textbf{K}$. The kernel $\textbf{K}$ regulates the frequency of the sine function. We add a constant value of 1 to break the sine function's rotation symmetry around the origin. We use a Sigmoid activation function to guarantee the output is between 0 and 1. The topology network can be formulated as follows:
\begin{equation}
T(\textbf{X}) = \rho_m= \sigma(\textbf{W}\sin(\textbf{K}\textbf{X} +  1))
\end{equation}

\noindent where:

$\textbf{X}$: Domain coordinate input, $\textbf{X}=(x,y,u,w)$ 

$\sigma$: Sigmoid activation function

$\textbf{K}$: Trainable frequency kernels

$\textbf{W}$: Trainable weights

\noindent 

Typically, the trainable kernel is initialized from a linear 4D mesh grid with frequency [-0.4,0.4] range for the ten local dimension kernels and frequency range [-0.6,0.6] for the six global dimension kernels. The weight is initialized to be a constant of 0.1 for problems except the mechanical cloaking to create some undulations in the micro design field. This yields 3.6k trainable frequency kernels and the same number of trainable weights. The input consists of a batch of coordinates. For a problem with no mini-epoch solving, the total batch size corresponds to the total number of elements $N_m$ in the multiscale structure. For regularization, we apply a small L2 loss on the trainable weights $\textbf{W}$. 

\subsection{Homogenization and Inverse Homogenization}
We rely on an energy-based homogenization approach to compute the elasticity tensor of each cell and design the material distribution of each cell. A detailed review and accompanying code can be found in the work from Xia and Breitkopf \cite{xia2015design}. Nevertheless, for the sake of clarity, we include a brief review of the formulation of homogenization and inverse homogenization in this section.  

Homogenization aims to evaluate the effective constitutive behavior of periodic microstructures. The energy-based approach, based on the average stress and strain theorems, is a compact and easy-to-implement formulation for homogenization. In this method, the unit cell is discretized into a finite element mesh, and the displacement field is calculated to correspond to a set of unit test strain fields under periodic boundary conditions. The homogenized stiffness tensor $E^H_{ij}$ (in Voigt notation) of a given microstructure can be expressed in terms of the mutual energy of discretized elements $Y$. 

\begin{equation}
E^H_{ij} = \frac{1}{|Y|}\sum_{e=1}^N (\textbf{u}_e^{ij})^T\textbf{k}_e \textbf{u}_e^{ij}
\end{equation}

\noindent where $|Y|$ is the area of the design domain, $N$ is the number of elements, $k_e$ is the element stiffness matrix, and $u_e^{ij}$ is the element displacement under the unit test strain fields $\varepsilon_0^{ij}$. In the inverse homogenization, we aim to find the density distribution $\rho_e$ within the unit cell region to achieve certain requirements on the stiffness tensor, which can be formulated as an optimization problem 

\begin{equation}
\begin{aligned}
    \min_{\rho_e} & \quad c_i\left(E^H_{ij}(\rho)\right) \\
    \text{s.t.} & \quad \mathbf{K} \mathbf{U}^{kl} = \mathbf{F}^{kl}, \quad k,l = 1, \dots, 3\\
                & \quad \sum_{e=1}^N v_e \rho_e / |Y| \leq V_e, \\
                & \quad 0 \leq \rho_e \leq 1, \quad e = 1, \ldots, N.
\end{aligned}
\end{equation}

The optimization equation involves $ \mathbf{K} $, the stiffness matrix, while $ \mathbf{U} $ and $ \mathbf{F}$ represent the displacement vector and the external force vector for the test case, respectively. Here $ v_e $ is the volume of an individual element, and $ V_e $ defines the upper limit for the volume fraction. The objective function $ c(E^H) $ depends on the homogenized stiffness tensors. For example, the maximization of the bulk modulus can be written with an objective function: 

\begin{equation}
c_i = -(E_{11} + E_{12} + E_{21} + E_{22})
\label{eq:bulk_modulus}
\end{equation}

The homogenized stiffness tensor is used in two ways in this work. For bulk modulus optimization under varying volume fraction, we directly use the stiffness tensor to formulate the structural loss function. The bulk modulus optimization is also included in the metamaterial design studies (metamaterial design, negative Poisson's ratio materials, and mechanical cloaking) for improving the mechanical performance. In the micro-scale optimization study, we use the homogenized stiffness tensor to assemble the macro stiffness matrix. Homogenization assumes scale separation and periodicity \cite{bens0e1988generating, michel1999fft,suzuki1991shape, groen2018homogenization}. For both use cases of homogenization, we ensure a smooth transition between cells through the boundary loss and extension of the finite element boundary into the neighboring cells; thus, homogenization still gives us acceptable accuracy. Related works on functionally graded material also support the use of homogenization \cite{li2018optimal,Stollberg2025Graded}.  In the results section, we conduct full-scale analyses to support this assumption.

\subsection{Micro Scale Topology Optimization}

We adopt the multiscale structural topology optimization formulation \cite{chen2024multi,xia2014concurrent,gao2019concurrent} and modified it to focus on only micro-scale optimization. We fix the macro scale design variables as ones and only optimize for the microstructure. We define a corresponding material microstructure $\Omega_m$ discretized into $N_m$ elements. The micro design variables (for each macro cell) are $\rho_m^j$ $(j = 1,2,\ldots,N_m)$. The optimization problem is stated as:

\begin{equation}
\begin{aligned}
&\text{Find: } \rho_m^j \quad (j = 1,2, \ldots, N_m) \\
&\text{Minimize: } c(\rho_m) \;=\; \frac{1}{2} \int_{\Omega_M} E_M(\rho_M, \rho_m)\,\varepsilon(\mathbf{u}_M)\,\varepsilon(\mathbf{u}_M)\,d\Omega_M \\
&\text{Subject to: } \\
&\quad a(\mathbf{u}_M, \boldsymbol{\nu}_M, E_M) \;=\; l(\boldsymbol{\nu}_M), 
\quad \forall\, \boldsymbol{\nu}_M \in H_{per}(\Omega_M, \mathbb{R}^d), \\
&\quad a(\mathbf{u}_m, \boldsymbol{\nu}_m, E_m) \;=\; l(\boldsymbol{\nu}_m), 
\quad \forall\, \boldsymbol{\nu}_m \in H_{per}(\Omega_m, \mathbb{R}^d), \\
&\quad V_m^*(\rho_m) \;=\; \int_{\Omega_m} \rho_m \, d\Omega_m \;-\; V_m \;\leq\; 0, \\
&\quad 0 < \rho_m^{\min} \;\leq\; \rho_m^j \;\leq\; 1.
\end{aligned}
\end{equation}

Here, $c(\rho_m)$ is the structural compliance. The macro topology is determined by $\rho_M$, and each macro element’s effective properties depend on its own micro topology described by $\rho_m$. $V_m^*(\rho_m)$ enforces a limit $V_m$ on the micro material consumption. The lower bounds $\rho_M^{\min}$ and $\rho_m^{\min}$ avoid numerical singularities. The displacement field at the macro scale is denoted by $\mathbf{u}_M$, with virtual displacement $\boldsymbol{\nu}_M \in H_{per}(\Omega_M, \mathbb{R}^d)$. The operator $a$ is a bilinear form representing the internal energy, and $l$ is a linear form associated with external forces. At the macro scale, the principle of virtual work leads to:

\begin{equation}
\begin{aligned}
\begin{cases}
  a(\mathbf{u}_M, \boldsymbol{\nu}_M, E_M) \;=\; 
      \displaystyle \int_{\Omega_M} E_M( \rho_m)\,\varepsilon(\mathbf{u}_M)\,\varepsilon(\boldsymbol{\nu}_M)\,d\Omega_M,\\[6pt]
  l(\boldsymbol{\nu}_M) \;=\; 
      \displaystyle \int_{\Omega_M} \mathbf{f}\,\boldsymbol{\nu}_M \, d\Omega_M \;+\; \int_{\Gamma_M} \mathbf{h}\,\boldsymbol{\nu}_M \, d\Gamma_M,\\[6pt]
  a(\mathbf{u}_m, \boldsymbol{\nu}_m, E_m) \;=\; 
      \displaystyle \int_{\Omega_m} E_m(\rho_m)\,\varepsilon(\mathbf{u}_m)\,\varepsilon(\boldsymbol{\nu}_m)\,d\Omega_m,\\[6pt]
  l(\boldsymbol{\nu}_m) \;=\; 
      \displaystyle \int_{\Omega_m} E_m(\rho_m)\,\varepsilon(\mathbf{u}_m^0)\,\varepsilon(\boldsymbol{\nu}_m)\,d\Omega_m.
\end{cases}
\end{aligned}
\end{equation}

In the above, $\mathbf{f}$ is the body force, and $\mathbf{h}$ is the boundary traction on the Neumann boundary $\Gamma_M$. The tensors $E_M(\rho_M,\rho_m)$ and $E_m(\rho_m)$ are derived via a material interpolation scheme akin to the modified Solid Isotropic Material with Penalization (SIMP) approach, given by:

\begin{equation}
  E_m \;=\; \Big[c_0 + (\rho_m)^p\bigl(1 - c_0\bigr)\Big]\;E_0,
\end{equation}

\noindent where $E_0$ is the base material elasticity tensor, $c_0 = 10^{-9}$ prevents numerical singularity, and $p$ is the penalization exponent.

A variety of gradient-based methods may be employed to solve this formulation. The sensitivities (first-order derivatives) of $c(\rho_m)$ and the volume constraints  $V_m^*(\rho_m)$ with respect to the design variables are needed. The partial derivatives of the objective and microstructure volume constraint with respect to $\rho_m$ read:

\begin{equation}
\begin{aligned}
\frac{\partial c}{\partial \rho_m}
\;=&\;
\frac{1}{2}\int_{\Omega_M}\Bigl[c_0 + (\rho_M)^p\bigl(1 - c_0\bigr)\Bigr]\\
&\times\frac{\partial\,E^H(\rho_m)}{\partial\,\rho_m}\varepsilon(\mathbf{u}_M)\,\varepsilon(\mathbf{u}_M)d\Omega_M,
\end{aligned}
\end{equation}

\noindent The derivative of the homogenized elastic tensor $E^H$ with respect to $\rho_m$ is given by:

\begin{equation}
\begin{aligned}
\frac{\partial E^H(\rho_m)}{\partial \rho_m}
\;=&\;
\frac{1}{|\Omega_m|}\,
\int_{\Omega_m}
p\,(\rho_m)^{\,p-1}\,\bigl(1 - c_0\bigr)\,
E_0\; \\
&\times \bigl[\varepsilon(\mathbf{u}_m^0) \;-\; \varepsilon(\mathbf{u}_m)\bigr]\,
\bigl[\varepsilon(\mathbf{u}_m^0) \;-\; \varepsilon(\mathbf{u}_m)\bigr]\,
d\Omega_m.
\end{aligned}
\end{equation}

In implementation, the changes occur in implementing a multiscale FE solver, which computes the compliance $c$ for the entire structure and adds the volume fraction constraint $V^*_M$ for the macro scale structure. The combined loss function for concurrent multiscale topology optimization is:

\begin{equation}
\mathcal{L} = \frac{c}{c_{0}} + \alpha(\frac{V_j}{V^*_m}-1)^2 + 0.1\alpha\mathcal{L}_{bc,i},
\end{equation}

\noindent where $c$ (without subscripts) is the compliance of the entire structure (not to be mistaken with the inverse homogenization objective function for each cell $c_i$). We compute the $c_0$ on the homogenized domain to normalize the compliance. $\alpha$ is a penalization value that gradually increases; compared to prior work that used a larger final value of $\alpha=100$, we reduced it to 50 such that the resulting metamaterial has slightly higher variability in its volume fraction. We also use $\alpha$ when enforcing the boundary loss $\mathcal{L}_{bc,i}$ with a reduced weight. The reduction is intended to roughly match the loss scale between the volume fraction and the boundary loss.  The boundary loss further enforces compatibility between cells.  A detailed visualization and explanation of the boundary loss can be found in \cite{chen2024multi}. 

\subsection{Metamaterial Design with Neural Networks}

The objective function first consists of a target displacement $\textbf{u}_t$ where the displacement $\textbf{u}$ of the multiscale structure is trying to match. Same with Wang et al. \cite{wang2020deep}, we minimize the difference in displacement with an L2 loss:
\begin{equation}
\mathcal{F} = ||\mathbf{\gamma}\circ\mathbf{u}-\mathbf{u}_t||^2_2
\end{equation}

\noindent Following Wang et al., we only design the microstructure in metamaterial system design. Thus, we modify their sensitivity function to compute the derivative of the objective function w.r.t micro design field $x_j$:

\begin{equation}
\begin{aligned}
\frac{\partial \mathcal{F}}{\partial x_j}
  &= -2 \bigl[\gamma \odot (u - u_t)\bigr]^T 
        K^{-1} \frac{\partial K}{\partial x_j} \,u,\\
&\quad \text{where}\quad 
  \frac{\partial K_e}{\partial x_j} 
    = B^T \frac{\partial C_e}{\partial x_j} \,B.
\end{aligned}
\end{equation}
%\frac{\partial \mathcal{F}}{\partial x_j}= -2 \bigl[\gamma \odot (u - u_t)\bigr]^T K^{-1} \frac{\partial K}{\partial x_j} \,u where\quad \frac{\partial K_e}{\partial x_j} = B^T \frac{\partial C_e}{\partial x_j} \,B

The combined loss function also resembles the previous examples of compliance minimization, but it includes a term for displacement mismatch and another term for bulk modulus maximization. We choose to add the bulk modulus maximization term in order to encourage connectivity and robustness of the designed structure for displacement matching. Otherwise, without this term, compared to data-driven methods with a library of feasible material, concurrent optimization may not prioritize overall structural stiffness. 

\begin{equation}
\mathcal{L} = \frac{1}{n}\sum^n \frac{c_i}{c_{0,i}} + \alpha||\mathbf{\gamma}\circ\mathbf{u}-\mathbf{u}_t||^2_2 + \alpha(\frac{V_i}{V^*_i}-1)^2 + 0.1\alpha\mathcal{L}_{bc,i}.
\end{equation}

\noindent Specifically for displacement matching, $\mathbf{\gamma}$ is a vector with the value one at the degrees of freedom corresponding to the displacement targets and with zeros everywhere else.

\section{RESULTS}
We set up metamaterial design problems with varying boundary conditions. These include metamaterial designs that target a specified displacement, NPR materials, and cloaking. We also include one micro-scale design problem with compliance minimization. These problems aim to test the ability of the neural network-based metamaterial to design compatible microstructures. All experiments are run on a PC with i7-12700K as the processor, 64 GB of RAM, and Nvidia RTX3080 GPU. 

\begin{figure*}
\centering
\begin{subfigure}[t]{0.3\textwidth}
\centering
\includegraphics[width=\textwidth]{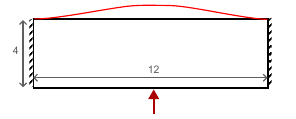}
\caption{}
\end{subfigure}
\qquad
\begin{subfigure}[t]{0.3\textwidth}
\centering
\includegraphics[width=\textwidth]{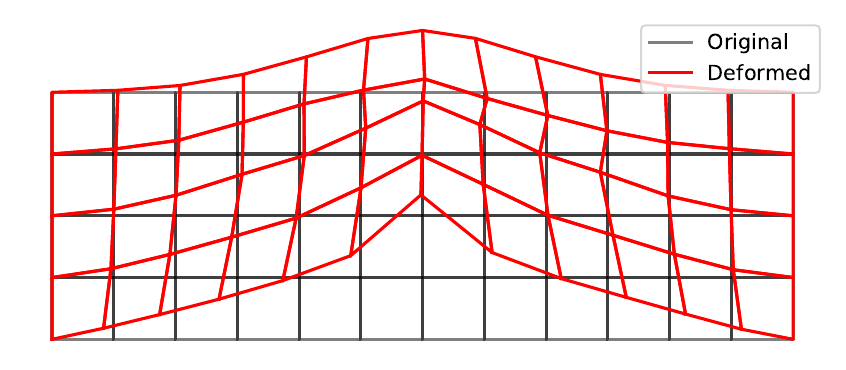}
\caption{}
\end{subfigure}

\begin{subfigure}[t]{0.3\textwidth}
\centering
\includegraphics[width=\textwidth]{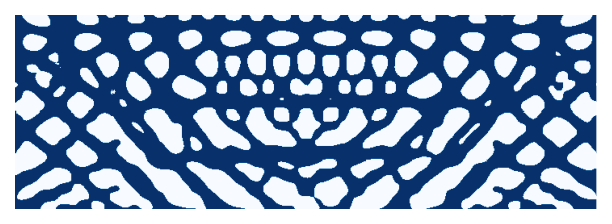}
\caption{}
\end{subfigure}
\qquad
\begin{subfigure}[t]{0.3\textwidth}
\centering
\includegraphics[width=\textwidth]{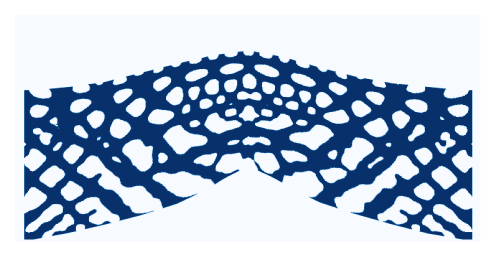}
\caption{}
\end{subfigure}

\caption{(a) The boundary condition setup for the metamaterial design problem, where the two sides are fixed, and a load is applied from the bottom center with the target displacement forming a bump at the top. (b) The deformed mesh corresponding to the optimized structure. (c) The converged geometry after 300 epochs of optimization using Adam with a learning rate of 0.001. (d) The deformed structure under the applied displacement field, demonstrating the connectivity and structural integrity of the optimized design. The Root Mean Square Error (RMSE) between the computed and target displacements is 0.556}
\label{fig:mm1x}
\end{figure*}

\begin{figure*}
\centering
\begin{subfigure}[t]{0.3\textwidth}
\centering
\includegraphics[width=\textwidth]{pic_mw1x_ms.pdf}
\caption{}
\end{subfigure}
\qquad
\begin{subfigure}[t]{0.3\textwidth}
\centering
\includegraphics[width=\textwidth]{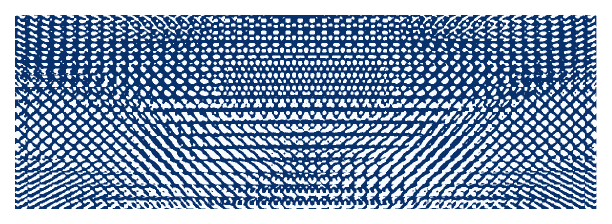}
\caption{}
\end{subfigure}

\begin{subfigure}[t]{0.3\textwidth}
\centering
\includegraphics[width=\textwidth]{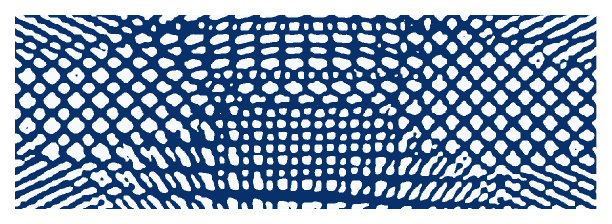}
\caption{}
\end{subfigure}
\qquad
\begin{subfigure}[t]{0.3\textwidth}
\centering
\includegraphics[width=\textwidth]{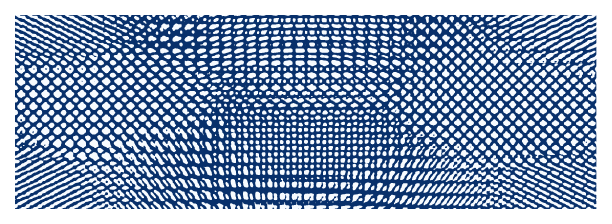}
\caption{}
\end{subfigure}

\begin{subfigure}[t]{0.3\textwidth}
\centering
\includegraphics[width=\textwidth]{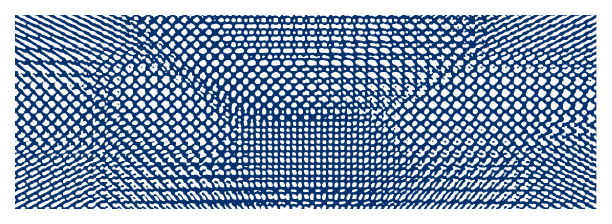}
\caption{}
\end{subfigure}
\qquad
\begin{subfigure}[t]{0.3\textwidth}
\centering
\includegraphics[width=\textwidth]{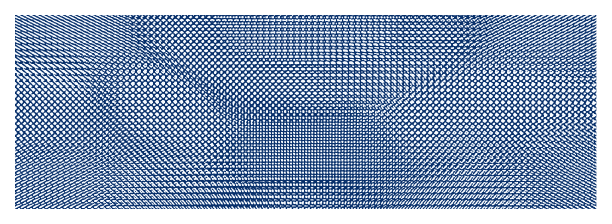}
\caption{}
\end{subfigure}

\begin{subfigure}[t]{0.6\textwidth}
\centering
\includegraphics[width=\textwidth]{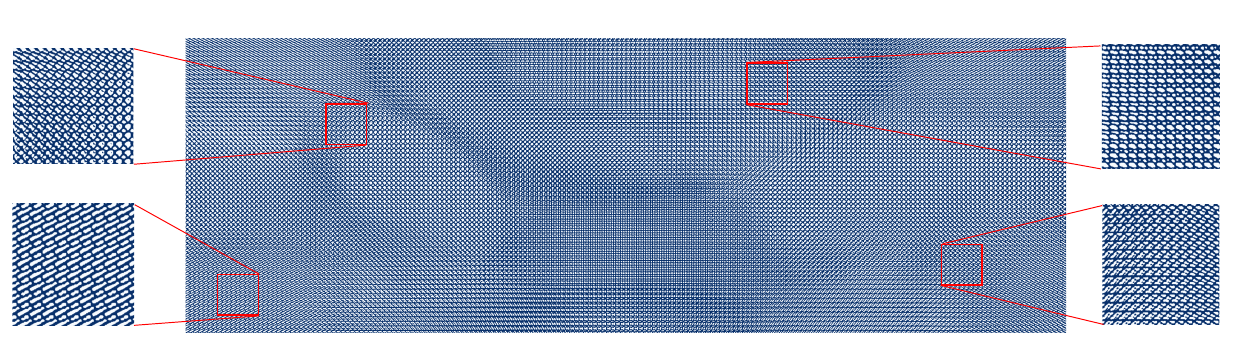}
\caption{}
\end{subfigure}

\caption{(a, c, e) The results of mini-epoch training with different global coordinate upsampling scales: no mini-epoch, 2× upsampling, and 4× upsampling, respectively. (b) The no-mini-epoch case upsampled post-training to 4× global coordinates, showing poor connectivity in the transition between microstructure cells. (d) The 2× upsampling case, further upsampled to 4× global coordinates after training, demonstrating improved connectivity. (f) The native 4× upsampled mini-epoch case, which exhibits the best connectivity among the tested configurations. A slight asymmetry is observed in the mini-epoch cases, likely due to the linear grid-like selection of representative cells during training, which does not preserve perfect symmetry across epochs. (g) Further upsampling of the 4$\times$ mini-epoch result to 16$\times$, we observe that good connectivity is still maintained.}
\label{fig:mm_nn}
\end{figure*}

\subsection{Metamaterial Design}
We first configure a simple metamaterial design to validate our method in generating connected multiscale structures and minimizing the objective function. The boundary condition is illustrated in Figure \ref{fig:mm1x} (a). We fix the two sides and apply a load from the bottom center. The target displacement forms a bump and is located at the top. We configured the optimization to run for 300 epochs with Adam as the optimizer and a learning rate of 0.001. 

After 300 epochs, we observe that the optimization has converged. We also plotted the deformed mesh in Figure \ref{fig:mm1x} (b) and applied the deformation to the structure in Figure \ref{fig:mm1x} (d). We evaluate the Root Mean Square Error (RMSE) between the computed and target displacements. The RMSE for this problem is 0.556. 

The global coordinates can be upsampled such that for every epoch, we choose one sub-cell to represent the stiffness of the entire cell. We refer to this as a mini-epoch, where traditionally, an epoch is defined as seeing the entire training dataset once, while in a mini-epoch, only a subset of the microstructure is seen by the optimizer. This can also be viewed as relying on the neural network's ability as a continuous function to interpolate between cells. This means that during an optimization step, NN updates its weights not only to optimize the sampled cell, but the effects of the updated weights will also propagate to its spatial neighbors and get them optimized. The mini-epoch solving does not significantly increase the runtime since the number of elements solved per epoch does not change. 

During mini-epoch training, we explore two different global coordinates upsampling scaling (2 and 4 times). The result for no mini-epoch, 2 and 4 times, is shown in Figure \ref{fig:mm_nn} (a,c,e). After training, we upsampled the global coordinates for the no mini-epoch and 2 times cases to 4 times. This aims to verify the continuity of the neural network field by increasing the scaling of the global coordinates. When upsampled to 4 times global coordinates, the no-mini epoch case (Figure \ref{fig:mm_nn} (b)) showed poor connectivity when inspecting the transition between cells. In the 2 times upsampling mini-epoch case, the 4 times upsampled result showed better connectivity. A native 4 times upsampled mini-epoch run is shown in Figure \ref{fig:mm_nn} (e), and it showed even better connectivity compared to the 2 times mini-epoch. One observation is that the mini-epoch case showed some slight asymmetry. We attribute this to the picking of the representative cells in each epoch following a linear grid-like fashion, which means that the selection process is not symmetrical, and across the epochs of optimization, the symmetry is not preserved. 

\begin{table}[h!]
\centering

\begin{tabular}{l|c|c|c}
%\hline
Global Upsampling & $1\times$ & $2\times$ & $4\times$ \\
\hline
Runtime & 12m06s & 12m40s & 12m50s \\

\end{tabular}
\caption{Runtime comparison}
\label{tab:mm_nn_runtime}
\end{table}

We summarize the runtime in Table \ref{tab:mm_nn_runtime}. We observe that there is only a slight increase in runtime when using the mini-epoch approach, even when we increase the global resolutions. For the 4 times upsampling result, we further increase the plotting resolution by showcasing 8 and 16 times upsampling in Figure \ref{fig:mm_nn} (f,g). At 16 times upsampling, the resulting field consists of over 44 million pixels, yet the overall connectivity is not broken during the upsampling process. When inspecting the transition between cells, there are cells with dangling/floating edges. We attribute this to an artifact of the continuous representation of the neural networks. Especially when an edge gradually disappears, the intermittent cells may show dangling or floating edges. While these are not common, morphological post-processing could be used to remove the floating edges.

\begin{figure*}
\centering
\begin{subfigure}[t]{0.4\textwidth}
\centering
\includegraphics[width=\textwidth]{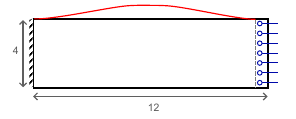}
\caption{}
\end{subfigure}
\qquad
\begin{subfigure}[t]{0.35\textwidth}
\centering
\includegraphics[width=\textwidth]{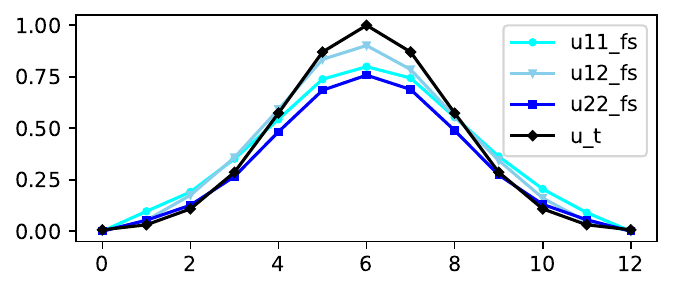}
\caption{}
\end{subfigure}

\begin{subfigure}[t]{0.25\textwidth}
\centering
\includegraphics[width=\textwidth]{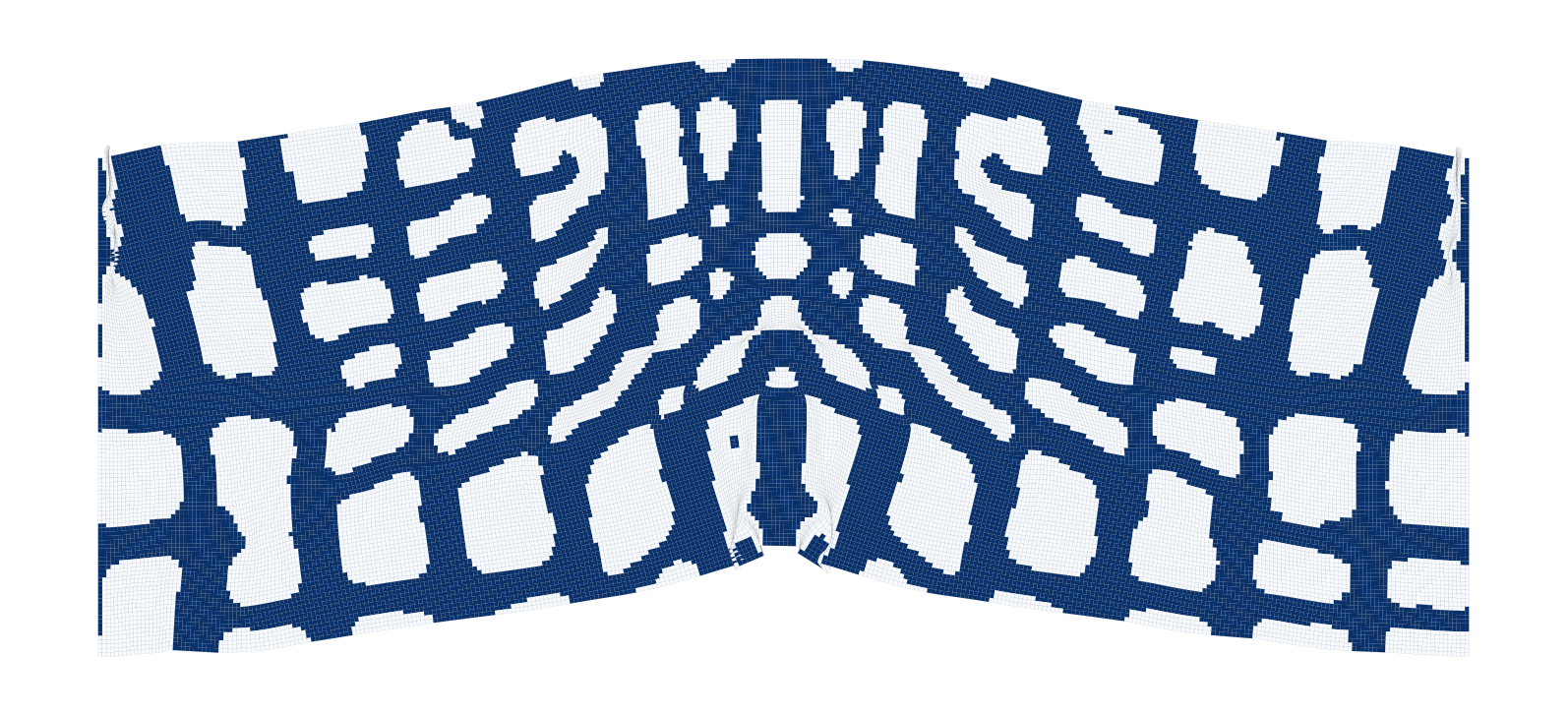}
\caption{}
\end{subfigure}
\qquad
\begin{subfigure}[t]{0.25\textwidth}
\centering
\includegraphics[width=\textwidth]{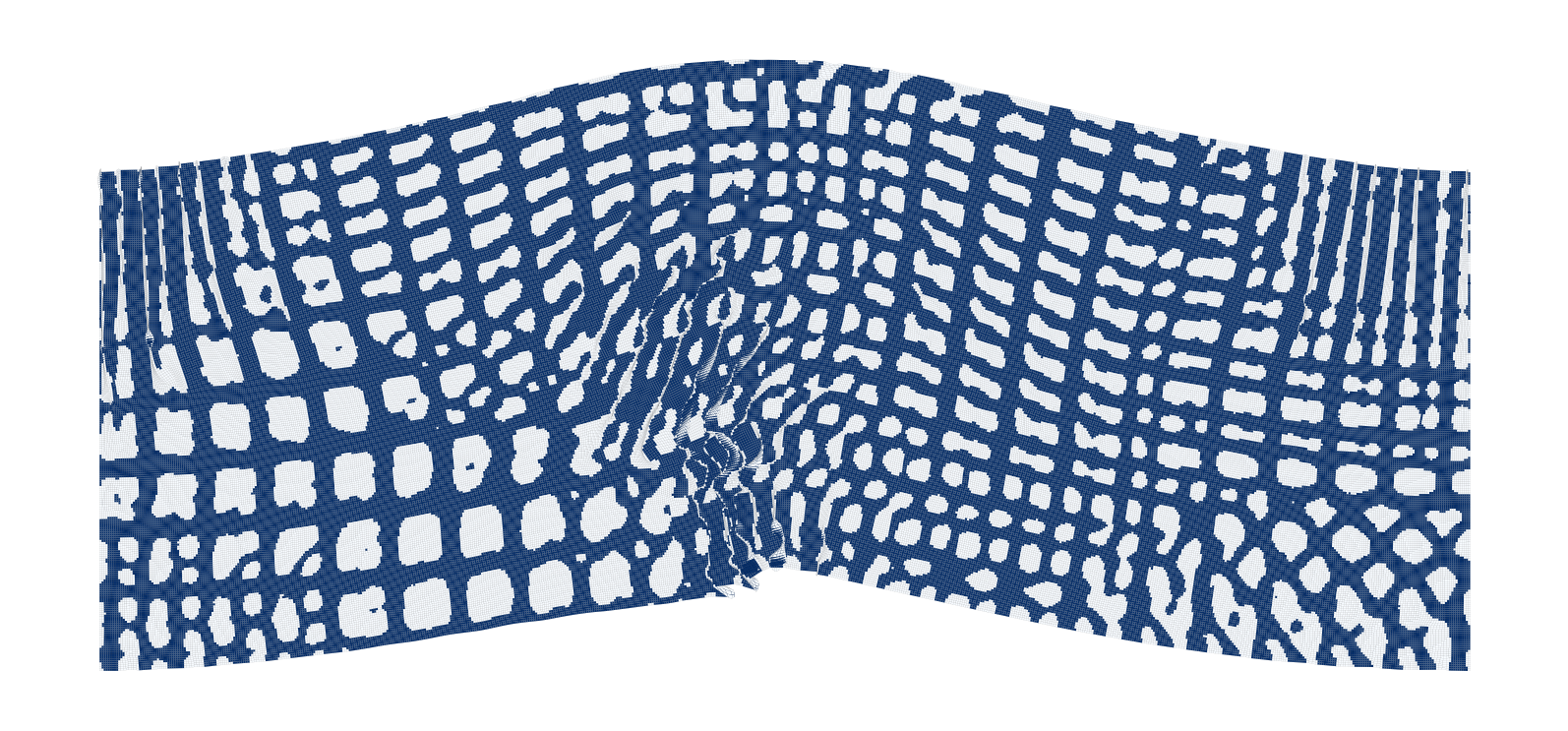}
\caption{}
\end{subfigure}
\qquad
\begin{subfigure}[t]{0.25\textwidth}
\centering
\includegraphics[width=\textwidth]{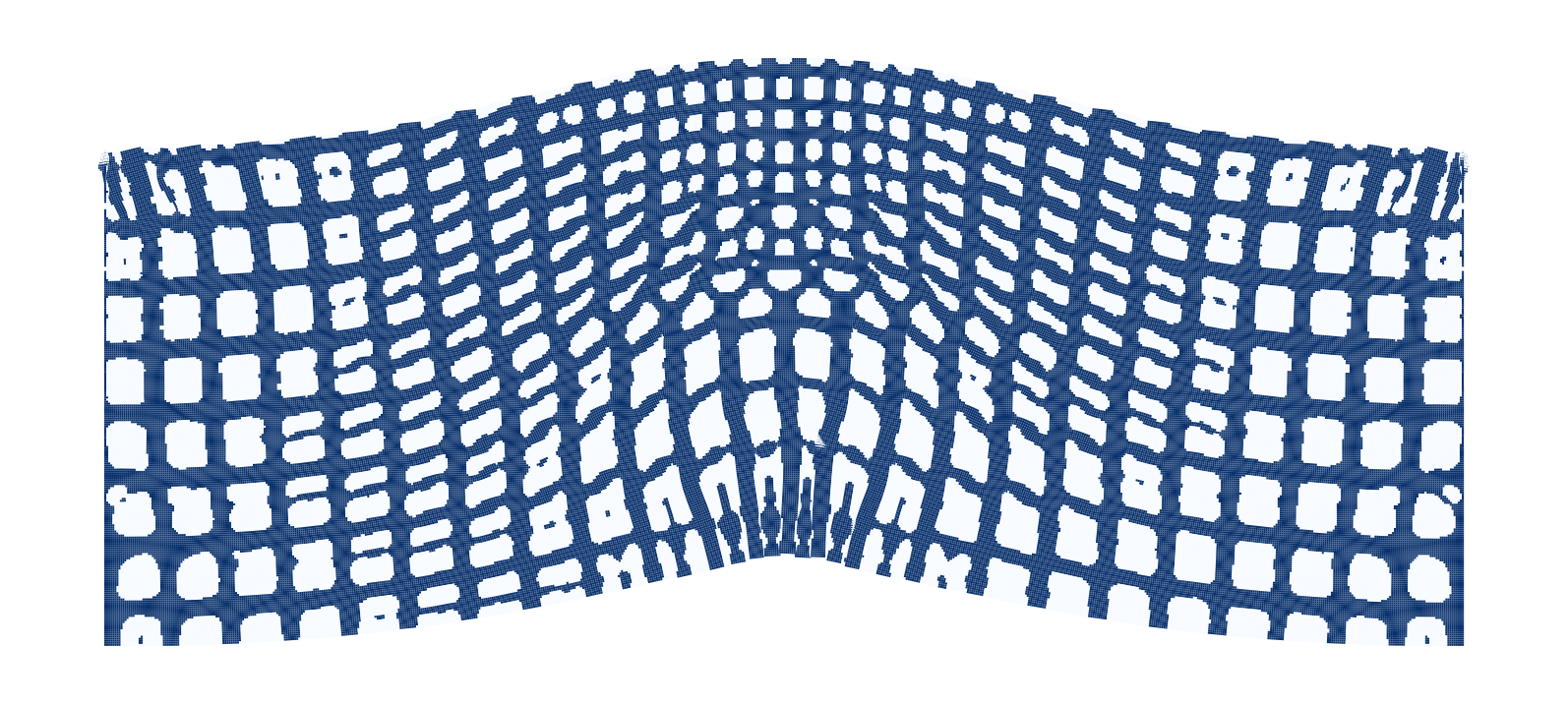}
\caption{}
\end{subfigure}

\caption{(a) The displacement boundary condition where we apply a unit displacement on the right-hand side. (b) Plotting out the full scale evaluated displacement with the target displacement. We observe that all full-scale analyses showed slightly smaller displacements than the target. (c) 12$\times$4 resolution without mini-epoch (u\_11), with a full-scale resolution of 360$\times$120. (d) 24$\times$8 resolution with mini-epoch (u\_12), with each batch consisting of 12$\times$4 macro cells and a full-scale resolution of 720$\times$240. (e) 24$\times$8 resolution without mini-epoch (u\_22), with a full-scale resolution of 720$\times$240. }
\label{fig:fs_disp}
\end{figure*}

\begin{table}[h!]
\centering
\scriptsize
\begin{tabular}{l|c|c}
%\hline
Configuration & Homogenized RMSE & Full-scale RMSE  \\
\hline
12$\times$4 & 0.003 & 0.450  \\
24$\times$8 w/ mini-epoch & 0.009 & 0.476  \\
24$\times$8 & 0.003 & 0.447  \\
\end{tabular}
\caption{Homogenized and full-scale RMSE comparison}
\label{tab:fs_homo_rmse}
\end{table}

The mini-epoch solving demonstrates good runtime reduction. We then set up a slightly modified problem with a displacement boundary condition (Figure \ref{fig:fs_disp}(a)) to run full-scale analysis on the metamaterial at the original and upsampled scales. We configure three runs. The original resolution run of 12$\times$4 without mini-epoch (Figure \ref{fig:fs_disp}(c)). A two-times upsampling mini-epoch run (Figure \ref{fig:fs_disp}(d)). And a run with native 24$\times$8 resolution run without mini-epoch (Figure \ref{fig:fs_disp}(e)). In (Figure \ref{fig:fs_disp}(b)), we show a full-scale evaluation of the displacement and target displacement comparison. We observe that despite the homogenized version closely matching the target displacement, all full-scale analyses showed slightly smaller displacements than the target. The detailed result is summarized in Table \ref{tab:fs_homo_rmse}, where all full-scale evaluations showed a small increase in the RMSE, with the native 24$\times$8 run showing the smallest full-scale error. Still, the full-scale analysis showed that the overall shape matches the target displacement. We suspect that for this specific problem, practitioners may increase the target displacement such that the full-scale analysis can better match the original target displacement. 

\begin{figure}
\centering
\begin{subfigure}[t]{0.45\textwidth}
\centering
\includegraphics[width=\textwidth]{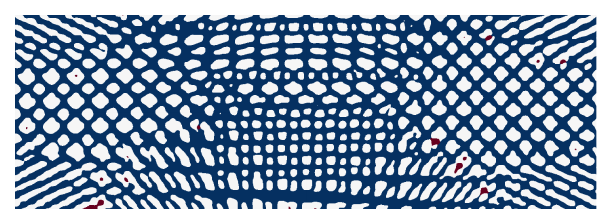}
\caption{}
\end{subfigure}

\begin{subfigure}[t]{0.45\textwidth}
\centering
\includegraphics[width=\textwidth]{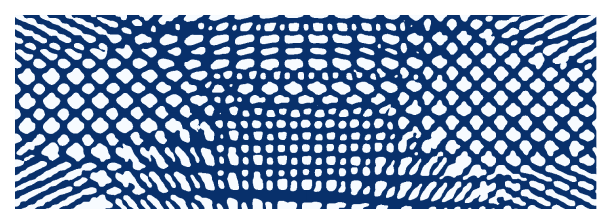}
\caption{}
\end{subfigure}

\caption{Island removal with thresholding. We first remove the dangling islands at a 0.3 cutoff. Then the cutoff is increased to 0.5, which causes some dangling edges to become a disconnected island. We then use these new disconnected islands as a mask to remove the dangling edges in the previous 0.3 cutoff. Here we use the result shown in Figure \ref{fig:mm_nn} (c) to demonstrate the post-processing. (a) Dangling edges and islands are detected and highlighted in red. (b) Result with dangling edges and islands removed. We note that some dangling edges remain, which would necessitate more comprehensive methods.   }
\label{fig:island}
\end{figure}

Upon closer inspection of the geometry, we noticed the formation of dangling edges and isolated islands during cell transitions. These artifacts are most pronounced when a cell begins forming a new edge based on its neighboring cells; in intermediate stages, this process can yield incomplete structures that become disconnected islands or dangling edges. To address this, we implement a thresholding-based post-processing method. First, we remove the most prominent islands using a cutoff of 0.3 with an area smaller than 400 pixels. We then increased the threshold to 0.5, which caused certain previously connected structures to become disconnected, effectively transforming some dangling edges into removable islands. These newly isolated regions were then used as a mask to eliminate the corresponding dangling edges in the 0.3 threshold result. We use the result in Figure \ref{fig:mm_nn} (c) as an example to remove dangling edges. In Figure \ref{fig:island}(a), we highlight the detected islands and dangling edges in red, and show the cleaned result in (b). While the approach improves the geometry, some dangling structures remain and would require more advanced morphological methods \cite{Sigmund2007}, void detection methods \cite{Liu2015, Gaynor2020}, or strain energy-based methods \cite{groen2018homogenization}.

\begin{figure}
\centering
\begin{subfigure}[t]{0.25\textwidth}
\centering
\includegraphics[width=\textwidth]{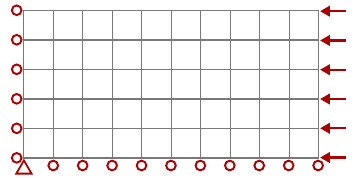}
\caption{}
\end{subfigure}

\begin{subfigure}[t]{0.25\textwidth}
\centering
\includegraphics[width=\textwidth]{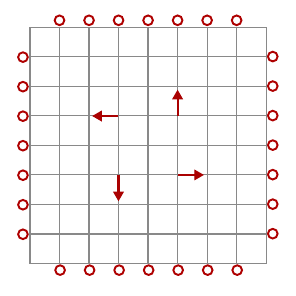}
\caption{}
\end{subfigure}

\caption{Boundary condition for the two NPR cases}
\label{fig:npr_bc}
\end{figure}

\begin{figure*}
\centering
\begin{subfigure}[t]{0.3\textwidth}
\centering
\includegraphics[width=\textwidth]{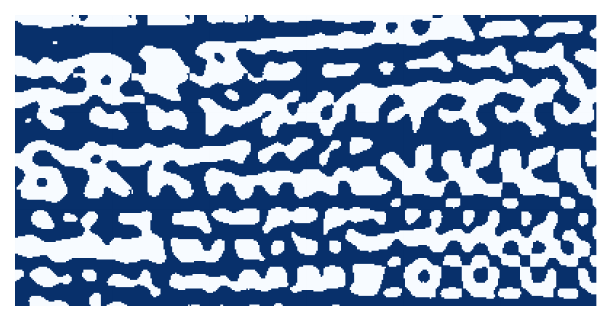}
\caption{}
\end{subfigure}
\qquad
\begin{subfigure}[t]{0.3\textwidth}
\centering
\includegraphics[width=\textwidth]{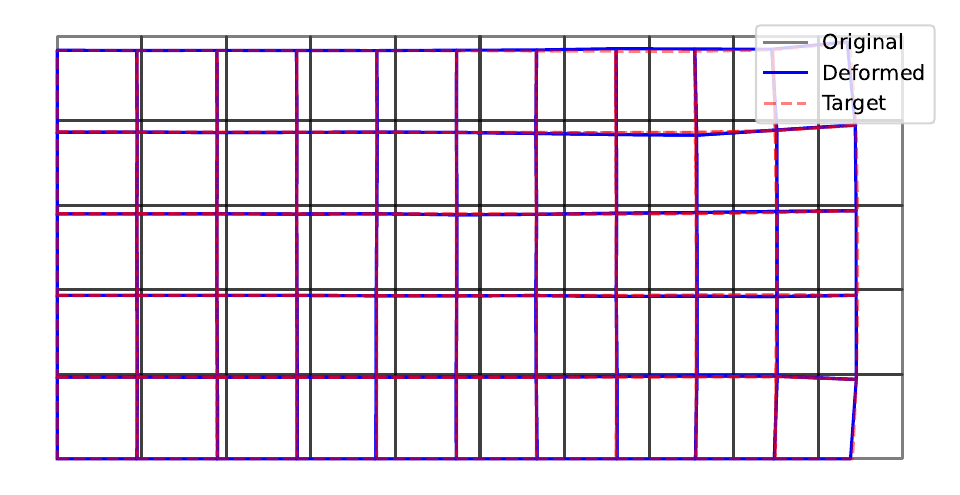}
\caption{}
\end{subfigure}
\qquad
\begin{subfigure}[t]{0.3\textwidth}
\centering
\includegraphics[width=\textwidth]{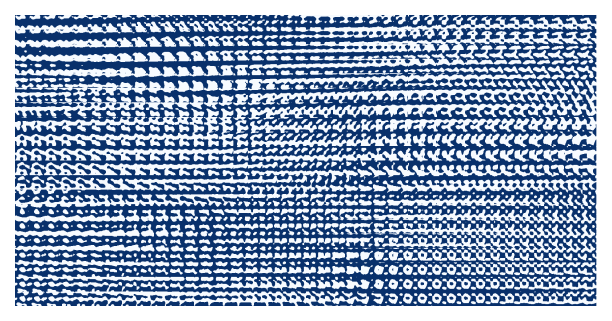}
\caption{}
\end{subfigure}

\begin{subfigure}[t]{0.3\textwidth}
\centering
\includegraphics[width=\textwidth]{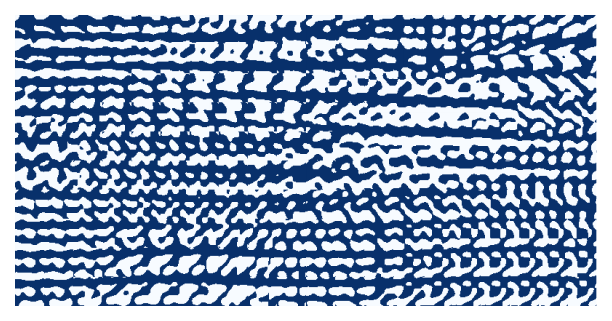}
\caption{}
\end{subfigure}
\qquad
\begin{subfigure}[t]{0.3\textwidth}
\centering
\includegraphics[width=\textwidth]{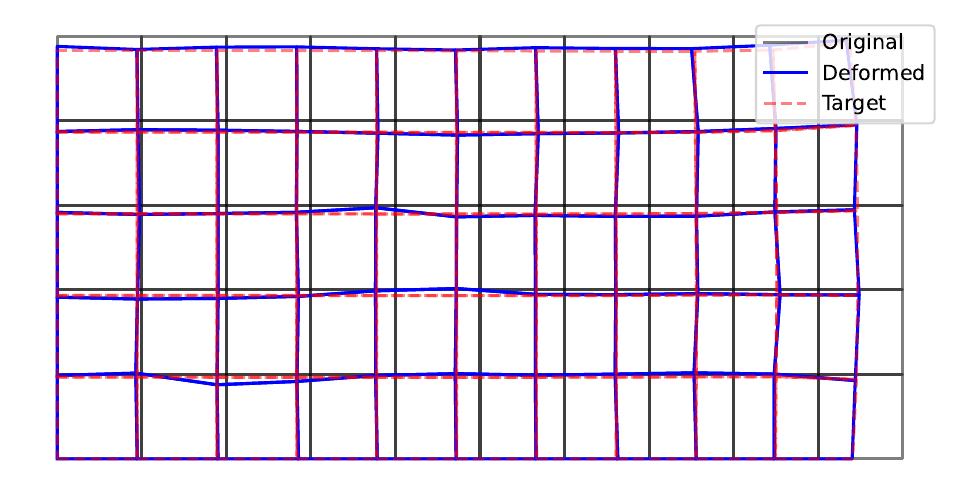}
\caption{}
\end{subfigure}
\qquad
\begin{subfigure}[t]{0.3\textwidth}
\centering
\includegraphics[width=\textwidth]{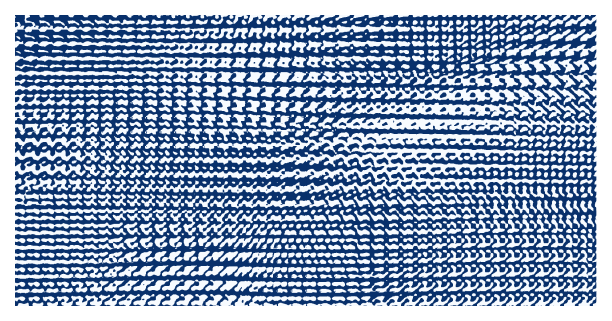}
\caption{}
\end{subfigure}

\caption{(a, d) The results for no mini-epoch and 2× upsampling mini-epoch cases. (b, e) The corresponding distorted meshes and target deformations illustrate the deformation patterns under the applied conditions. The no-mini-epoch case exhibits noticeable discontinuities between microstructure cells.}
\label{fig:npr_nn}
\end{figure*}

\begin{figure*}
\centering
\begin{subfigure}[t]{0.3\textwidth}
\centering
\includegraphics[width=\textwidth]{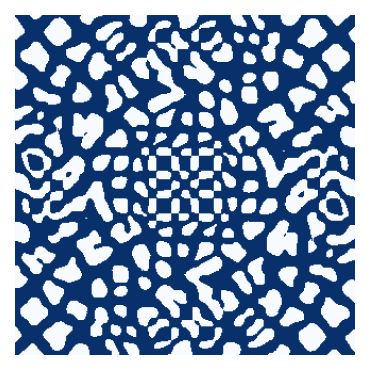}
\caption{}
\end{subfigure}
\qquad
\begin{subfigure}[t]{0.3\textwidth}
\centering
\includegraphics[width=\textwidth]{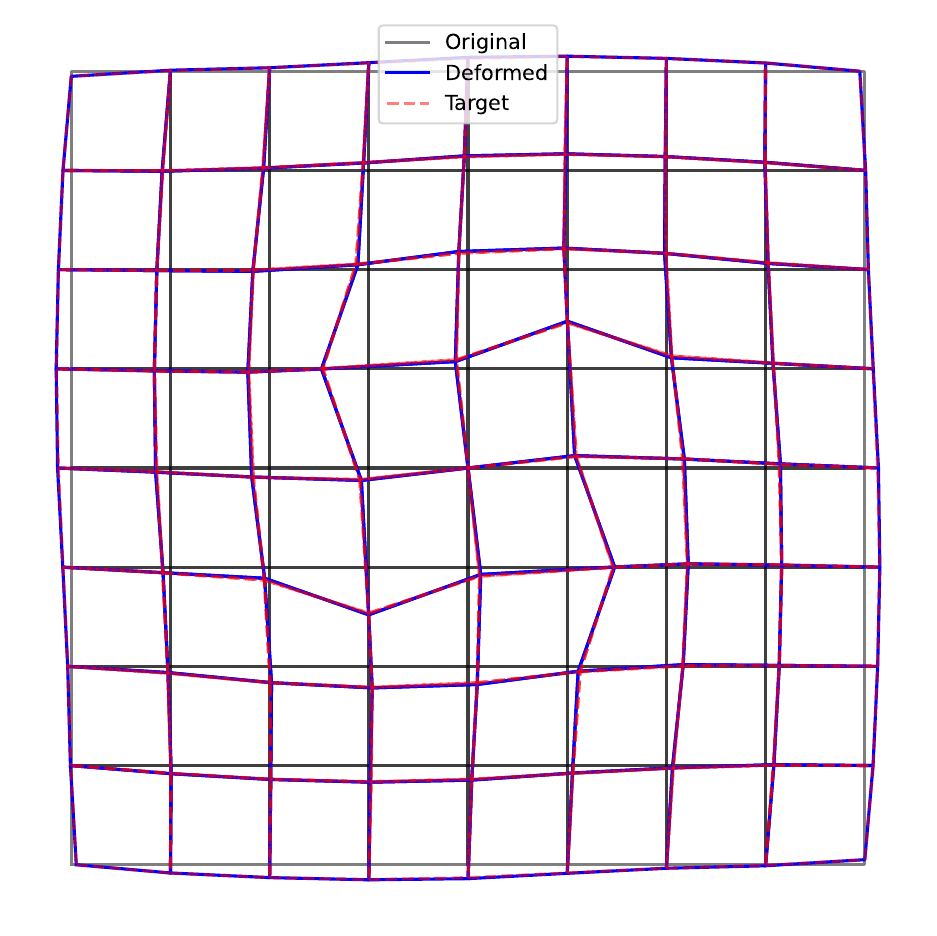}
\caption{}
\end{subfigure}
\qquad
\begin{subfigure}[t]{0.3\textwidth}
\centering
\includegraphics[width=\textwidth]{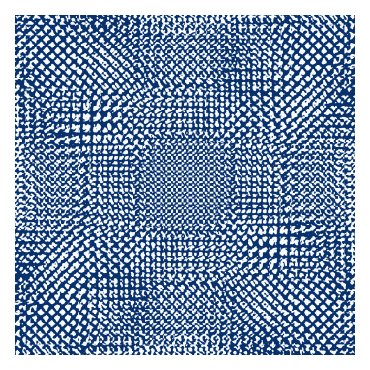}
\caption{}
\end{subfigure}

\begin{subfigure}[t]{0.3\textwidth}
\centering
\includegraphics[width=\textwidth]{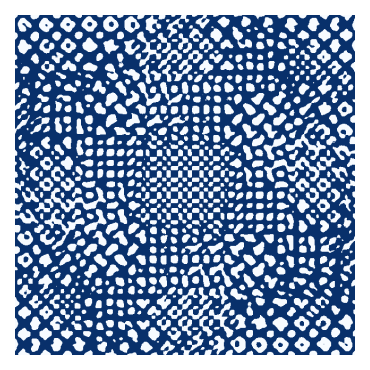}
\caption{}
\end{subfigure}
\qquad
\begin{subfigure}[t]{0.3\textwidth}
\centering
\includegraphics[width=\textwidth]{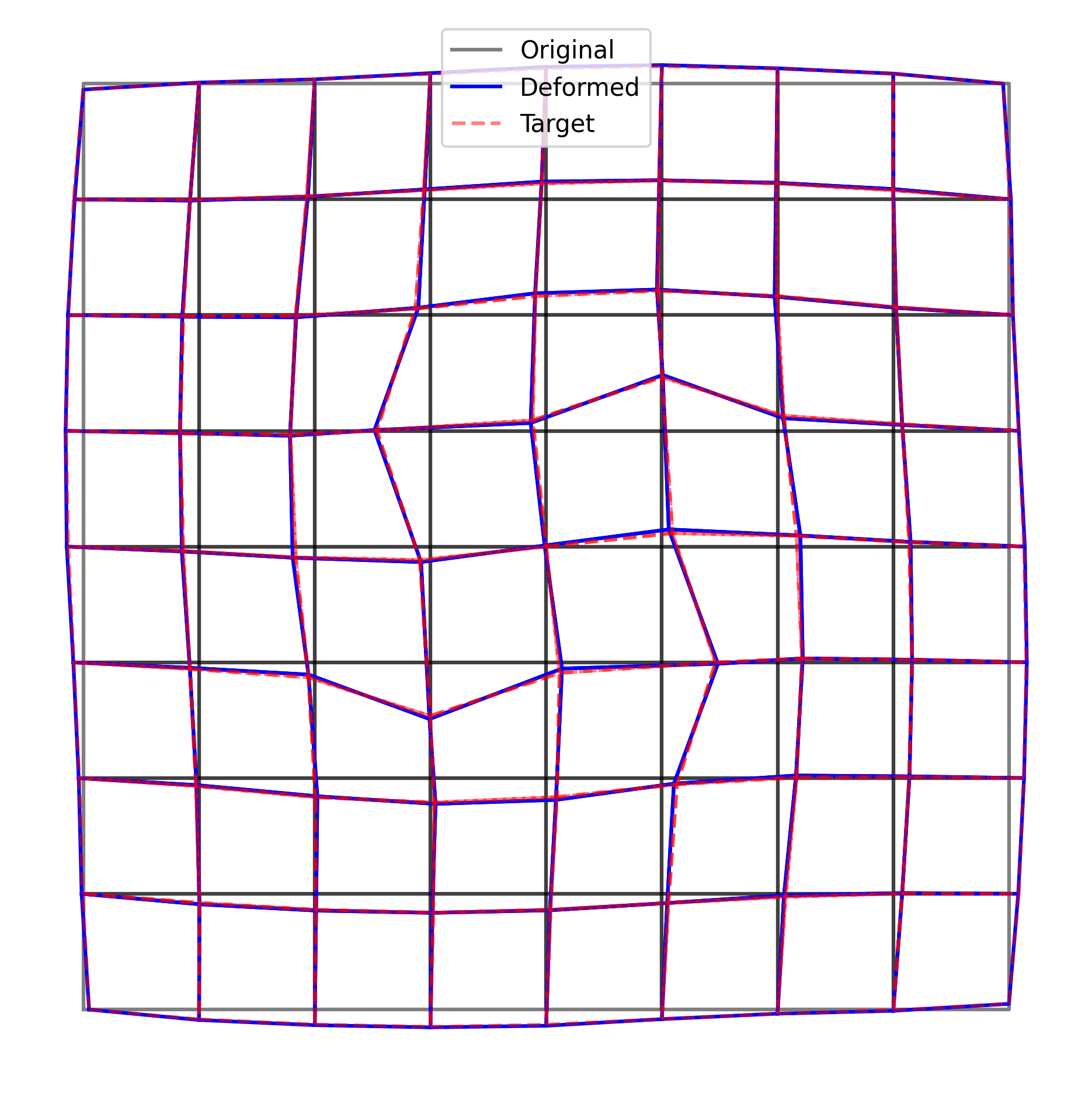}
\caption{}
\end{subfigure}
\qquad
\begin{subfigure}[t]{0.3\textwidth}
\centering
\includegraphics[width=\textwidth]{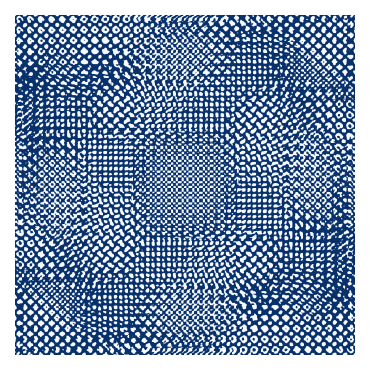}
\caption{}
\end{subfigure}

\caption{ (a, d) The same optimization configurations but evaluated under a different problem setup. (b, e) The distorted meshes and target deformations for these cases. (c, f) The results after upsampling to 4× resolution to assess the continuity of the neural network field. The 2× upsampling mini-epoch case demonstrates improved cell compatibility compared to the no-mini-epoch case, similar to previous findings.}
\label{fig:nprs_nn}
\end{figure*}

\subsection{Negative Poisson's Ratio Material}
NPR materials are materials that shrink along one direction when subjected to pressure along an orthogonal direction \cite{fan2021review}. We tested our method by generating NPR material. We first simulate the mechanical response under the boundary condition with a hypothetical material with a Poisson's ratio of -0.3 and record the displacement as the target. The two boundary conditions are illustrated in Figure \ref{fig:npr_bc}. We then run both of the experiments at no mini-epoch and 2x upsampling mini-epoch. The results are shown in Figure \ref{fig:npr_nn} (a,d) and Figure \ref{fig:nprs_nn} (a,d). 

\begin{table}[h!]
\centering

\begin{tabular}{l|c|c}
%\hline
Global Upsampling & $1\times$ & $2\times$  \\
\hline
RMSE 1 & 0.009 & 0.024  \\
RMSE 2 & 0.007 & 0.011  \\
\end{tabular}
\caption{RMSE comparison from the homogenized result }
\label{tab:npr_rmse}
\end{table}

We also evaluated the RMSE for both of the test cases summarized in Table \ref{tab:npr_rmse}. We observe that the mini-epoch does see a slight increase in an overall small RMSE error. Inspecting the distorted mesh and the target deformation as shown in the center image of Figure \ref{fig:npr_nn} (b,e) and Figure \ref{fig:nprs_nn} (b,e). We also upsampled the result to 4 times the resolution to compare the continuity of the neural network field. Similar to the previous example, we also observe the 2 times upsampled mini-epoch has better cell compatibility. 

\begin{figure*}
\centering
\begin{subfigure}[t]{0.34\textwidth}
\centering
\includegraphics[width=\textwidth]{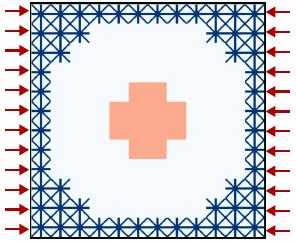}
\caption{}
\end{subfigure}
\qquad
\begin{subfigure}[t]{0.3\textwidth}
\centering
\includegraphics[width=\textwidth]{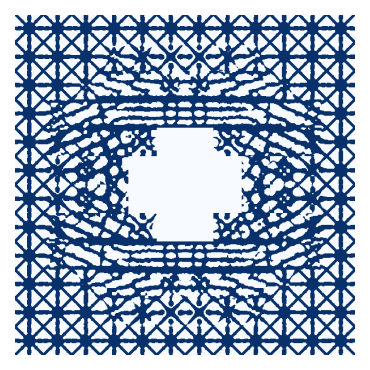}
\caption{}
\end{subfigure}

\begin{subfigure}[t]{0.3\textwidth}
\centering
\includegraphics[width=\textwidth]{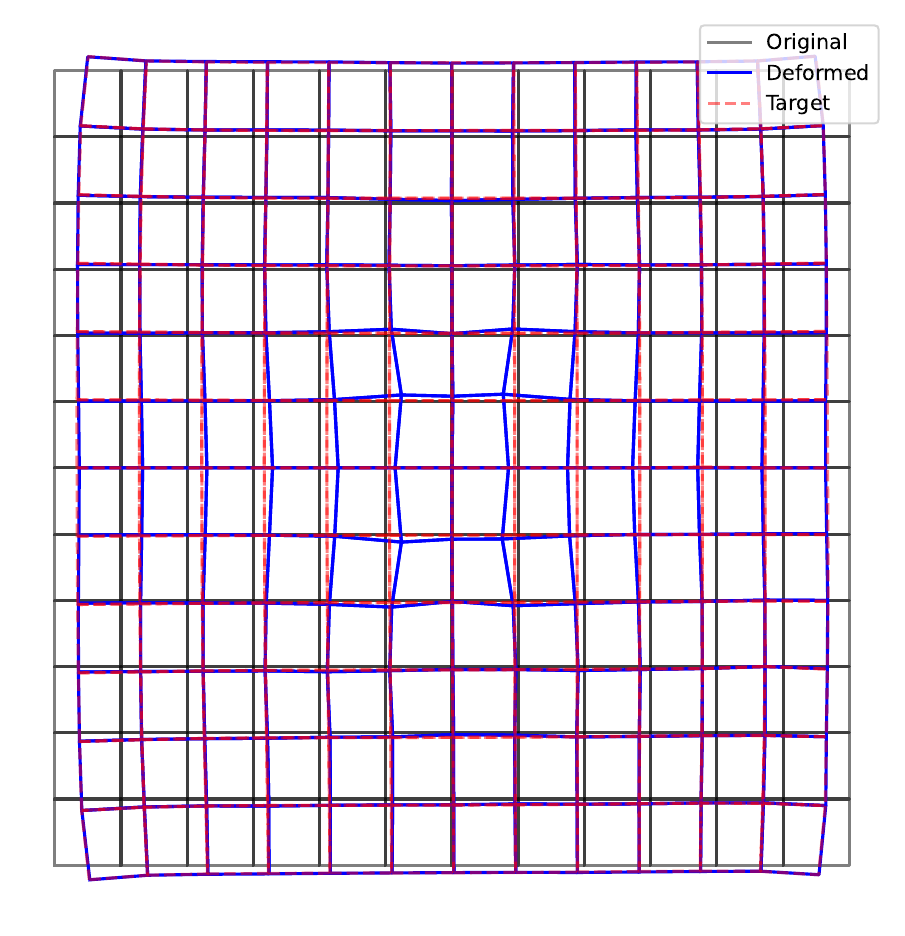}
\caption{}
\end{subfigure}
\qquad
\begin{subfigure}[t]{0.3\textwidth}
\centering
\includegraphics[width=\textwidth]{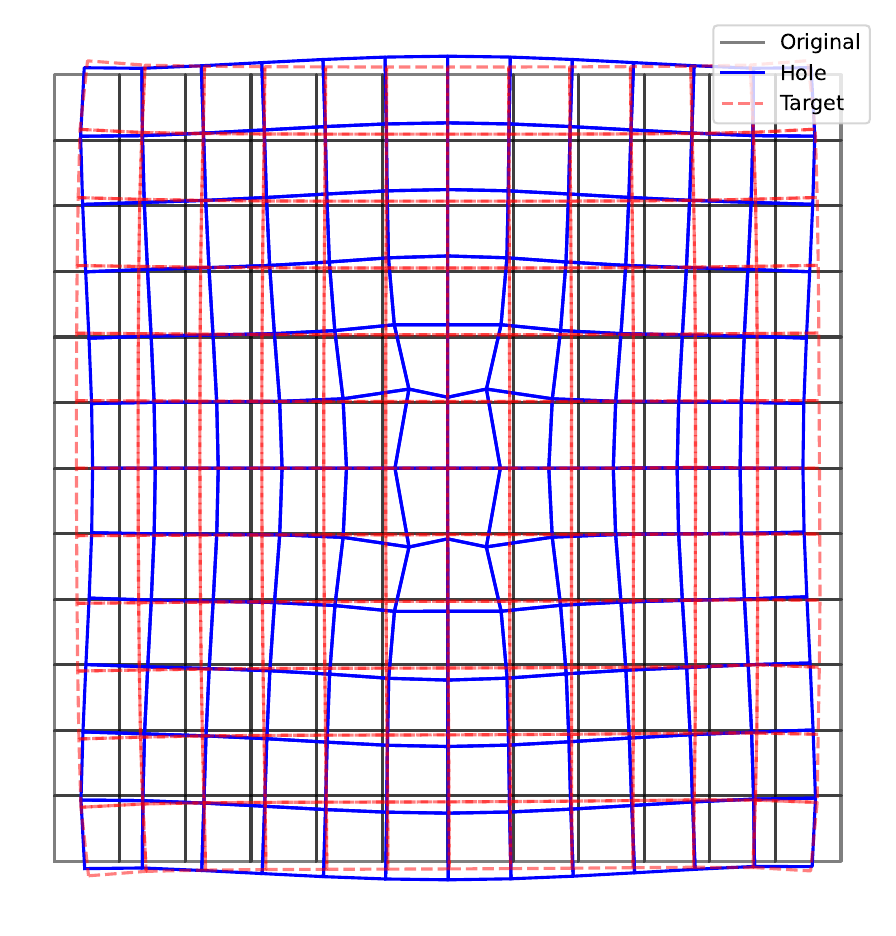}
\caption{}
\end{subfigure}

\caption{(a) The boundary condition, base cell, and hole configuration used in the optimization process. L1 loss is applied to enforce similarity between the neural network output and the base cell along the boundary. (b) The optimized result, where additional supporting structures form around the hole to bridge the gap. The smooth transition between the base cell and newly generated cells is observed, with partial resemblance to the base cell even beyond the enforced boundary region. (c) The deformed mesh visualized alongside the target displacement, with an RMSE of 0.011, demonstrating the effectiveness of the optimization. (d) The displacement field evaluation for the base cell with a hole, yielding an RMSE of 0.1}
\label{fig:cloak}
\end{figure*}

\begin{figure*}
\centering

\includegraphics[width=0.8\textwidth]{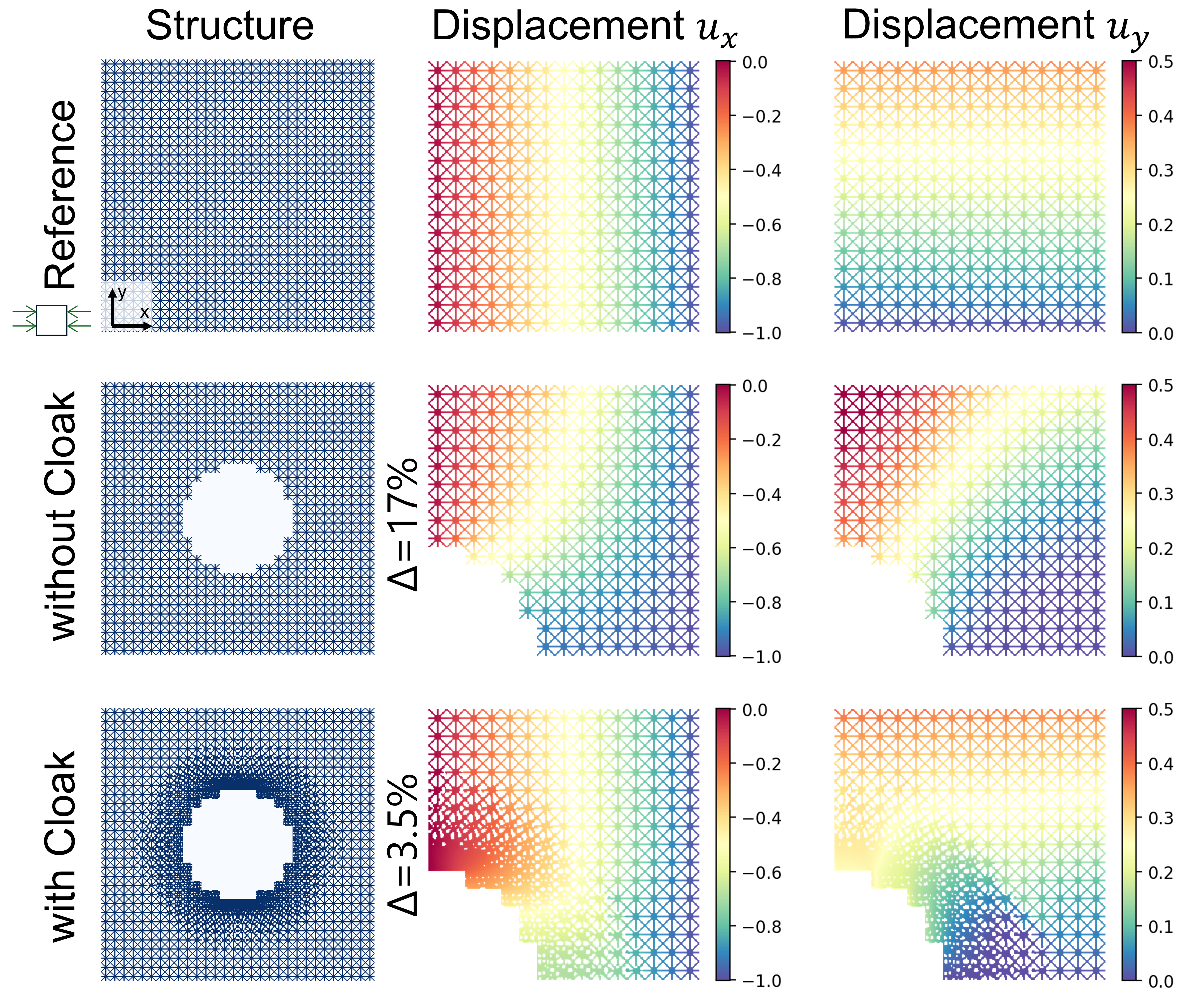}

\caption{Recreating the mechanical cloak example by Wang et al.'s \cite{wang2022mechanical}. We reduced the cell resolution by half while keeping the same displacement boundary condition, design domain, and setup for full-scale finite element analysis. We observe a slightly smaller $\Delta$ value of 3.5\% compared to the 4.0\% reported in Wang et al.'s paper. Similar to the previous smaller resolution example, the transition between the base cell and the designed cell is smooth and gradual.  }
\label{fig:cloak_bm}
\end{figure*}

\subsection{Mechanical Cloak}
Mechanical cloaks are specially designed materials that control the elastic behavior around objects, making these objects appear identical to their uniform environment \cite{wang2022mechanical}. Following the work of mechanical cloak via data-driven metamaterial by Wang et al., we aim to recreate the experiment with our neural network. We use the same base cell and create a hole around the center. The boundary condition, base cell, and hole are illustrated in Figure \ref{fig:cloak} (a). We then add an L1 loss to enforce the neural network to output similar cells to the base cell along the boundary. We compute the target displacement with the structure filled with the base cell. Next we optimize for both the resemblance of the base cell around the corner and minimizing the displacement mismatch between the outer base cells. The converged result is shown in Figure \ref{fig:cloak} (b), where we can observe due to the hole in the middle, more supporting structures are formed along it to enhance the structural integrity. Since the neural network was trained to mimic the base cell around the outer corner, the transition between the base cell and newly generated cells is smooth. We observe cells that partially resemble the base cell even outside the corner boundary. We then visualize the deformed mesh and compare it against the target displacement in Figure \ref{fig:cloak} (c) with an RMSE of 0.011. For the base cell with a hole, we also evaluate the displacement in Figure \ref{fig:cloak} (d), where the RMSE is 0.1, which shows the efficacy of our method.

We recreate the example by Wang et al. \cite{wang2022mechanical} with the same boundary condition, design domain, and base cell shape, while reducing the cell resolution by half for computation efficiency. The cloaked result achieved a slightly smaller $\Delta$ value of 3.5\% compared to the 4\% reported in the prior work. The $\Delta$ is computed as: 

\begin{equation}
\Delta = \frac{\sqrt{\sum(u-u_t)^2}}{\sum{(u_t)}^2}
\end{equation}

Inspecting the cell design, we observe that the transition between base cells and optimized cells is gradual and continuous.

\begin{figure*}
\centering
\begin{subfigure}[t]{0.18\textwidth}
\centering
\includegraphics[width=\textwidth]{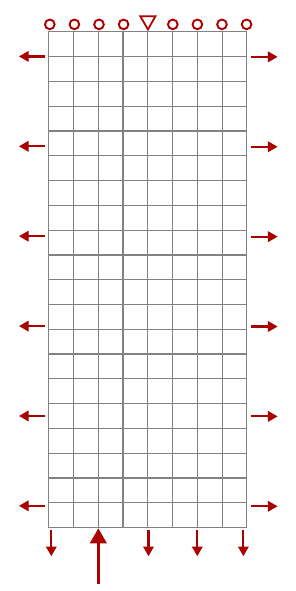}
\caption{}
\end{subfigure}
\qquad
\begin{subfigure}[t]{0.15\textwidth}
\centering
\includegraphics[width=\textwidth]{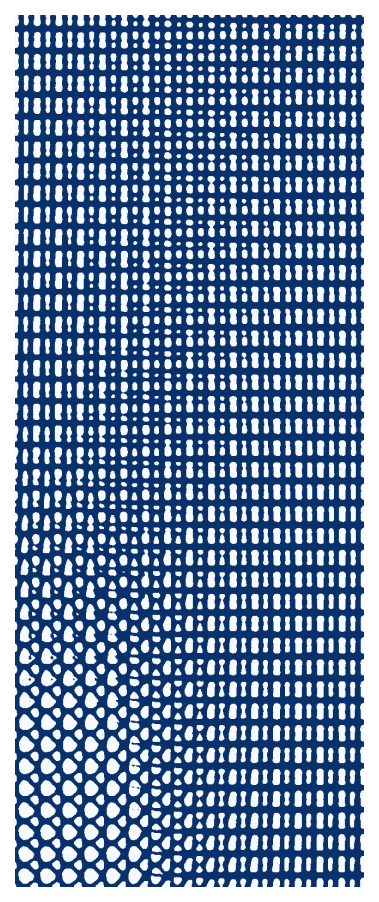}
\caption{}
\end{subfigure}
\qquad
\begin{subfigure}[t]{0.15\textwidth}
\centering
\includegraphics[width=\textwidth]{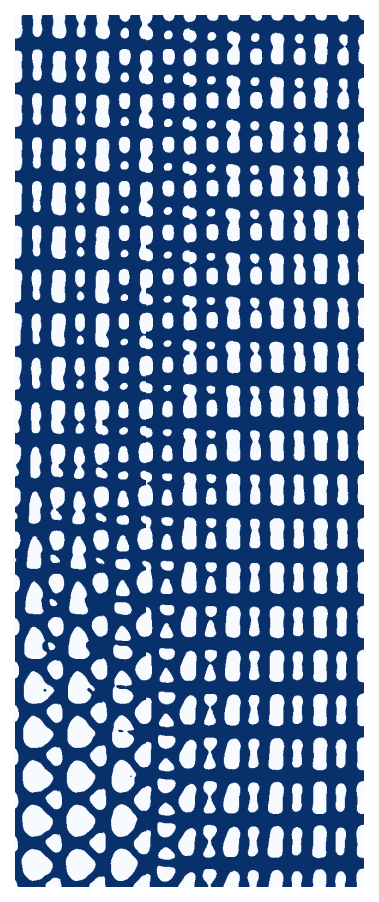}
\caption{}
\end{subfigure}
\qquad
\begin{subfigure}[t]{0.15\textwidth}
\centering
\includegraphics[width=\textwidth]{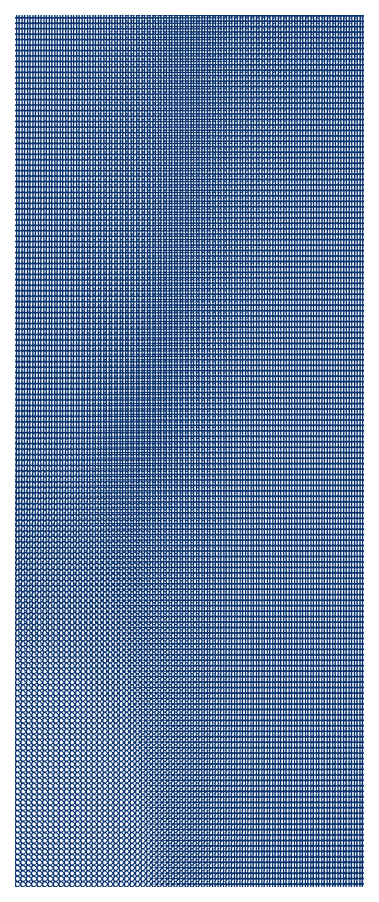}
\caption{}
\end{subfigure}

\begin{subfigure}[t]{0.5\textwidth}
\centering
\includegraphics[width=\textwidth]{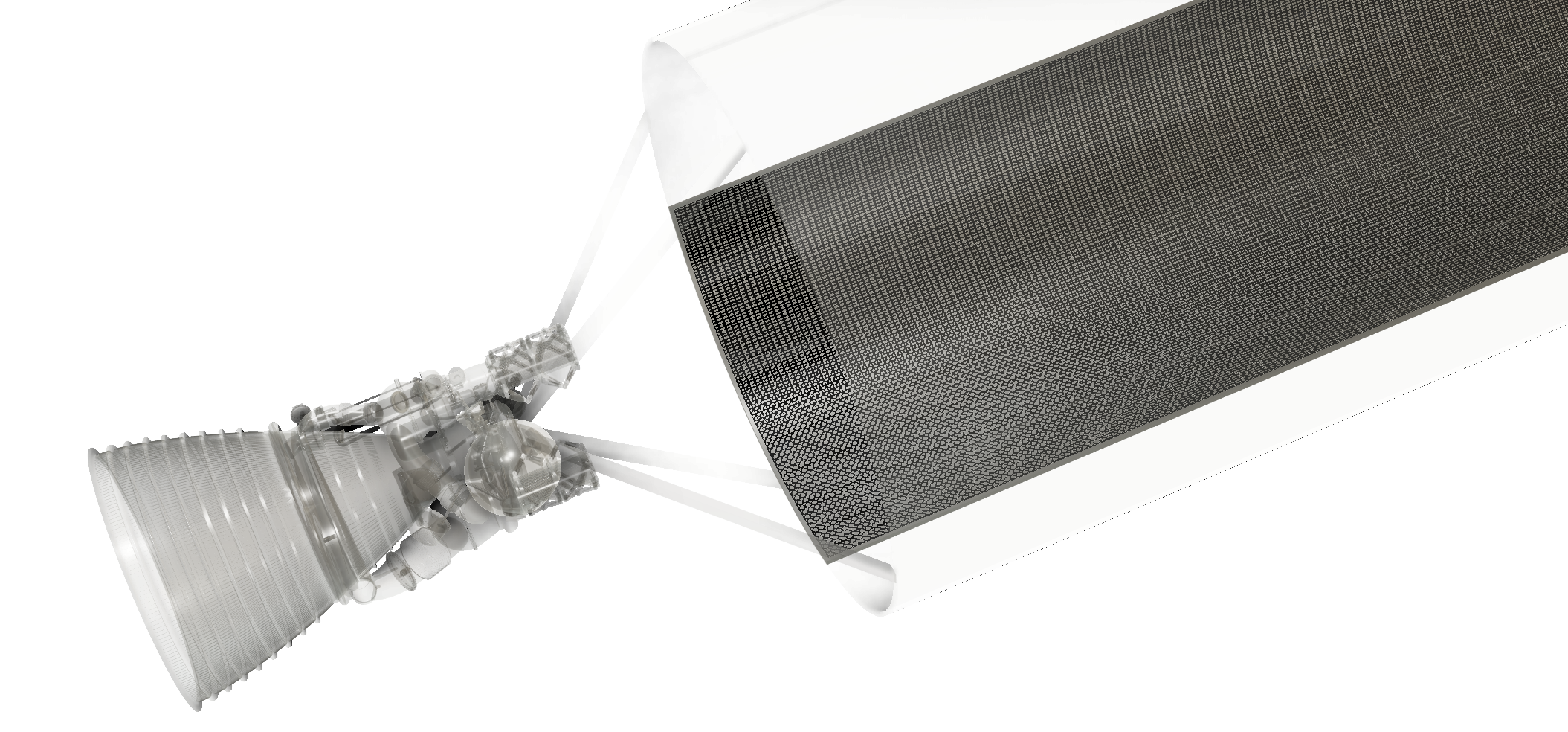}
\caption{}
\end{subfigure}

\caption{(a) The boundary conditions for optimizing grid structures within a rocket tank under pressurized propellant load and engine forces. (b) The optimized structure using a 2× upsampling mini-epoch, where rectangular grids effectively distribute the load across most of the tank, while shear-resistant structures emerge around the engine mount due to force concentration. (c) The 2× upsampled result downsampled to the original resolution, demonstrating that connectivity between cells is maintained. (d) The structure further upsampled to 8× resolution, maintaining the intended design features. (e) The 8× upsampled result applied as a texture on a rocket stage model with an attached engine. (Engine model by Grimmdp on Thingiverse under CC BY-NC license.)}
\label{fig:tank}
\end{figure*}
\subsection{Light-Weight Design}
Structural optimization with compliance minimization for lightweighting is also considered a type of metamaterial design \cite{yu2018mechanical}. We explore the design of grid structures within a rocket tank under a load of pressurized propellant and the force of the engine. The boundary condition is shown in Figure \ref{fig:tank} (a). We run the problem at 2 times upsampling mini-epoch. The result is shown in Figure \ref{fig:tank} (b). The result shows that around most of the tank structures, rectangular grids are sufficient to distribute the load. While around the force concentration of the engine mount, more shear-resistant structures are desired. We further explore whether the upsampled mini-epoch result can be down-sampled while maintaining the connectivity. We down-sampled the 2 times result to the original resolution as shown in Figure \ref{fig:tank} (c). We observe that with downsampling, there do not seem to be issues with cell compatibility. We then upsampled the result to 8 times resolution in Figure \ref{fig:tank} (d) and applied it as a texture to show the effect of the design on a rocket stage example with an engine attached (Figure \ref{fig:tank} (e)). 

\subsection{Bulk Modulus Optimization}

\begin{figure}
\centering
\begin{subfigure}[t]{0.45\textwidth}
\centering
\includegraphics[width=\textwidth]{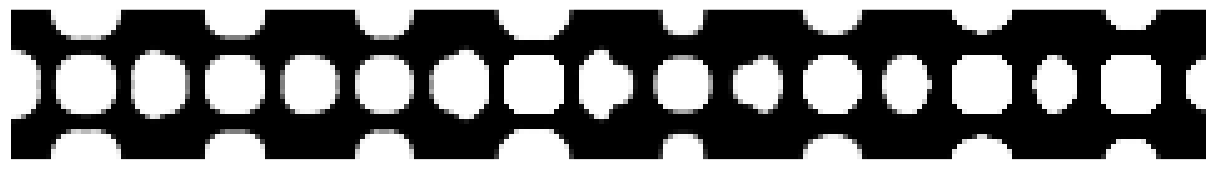}
\caption{}
\end{subfigure}

\begin{subfigure}[t]{0.45\textwidth}
\centering
\includegraphics[width=\textwidth]{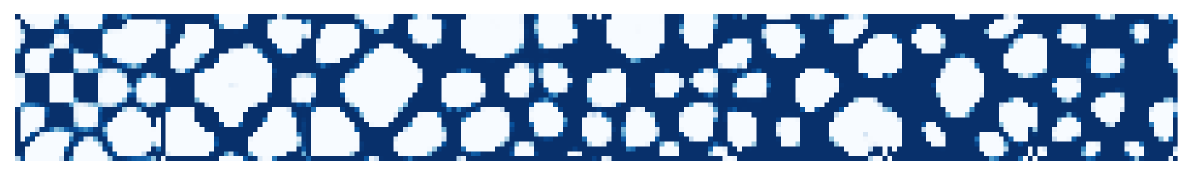}
\caption{}
\end{subfigure}

\caption{Bulk modulus optimization with (a) the MATLAB code by Garner et al. and (b) our metamaterial design framework}
\label{fig:bulkm}
\end{figure}

Bulk modulus is a measure of a material's resistance to uniform compression, quantifying how much it resists volume change under applied pressure. We configure a simple test case of bulk modulus maximization with our metamaterial design framework. Bulk modulus optimization also allows us to compare to the theoretical Hashin-Shtrikman (HS) upper bound \cite{hashin1963variational} of bulk modulus given a fixed volume fraction. We also report on the runtime with our method and a MATLAB code by Garner et al. \cite{garner2019compatibility}. For the best compatibility with the published code by Gerner et al., we configure a simple 8$\times$1 macro scale and 30$\times$30 micro scale with volume fraction linearly increasing from 0.4 to 0.7 and cell compatibility enabled. We use the function call: \texttt{topX\_Dual(8,30,0.4,0.7,0,1,1,0,0,1,1000)} and run both on the same PC with the same specifications. 

\begin{table}[h!]
\centering
\scriptsize
\begin{tabular}{l|c|c|c}
%\hline
 & Per cell \%HS & Avg & Runtime (s)  \\
\hline
Garner et al. & 74 72 72 81 72 80 83 85 & 77.4 & 2362  \\
Ours & 72 78 85 84 85 85 85 87  & 82.5 & 74  \\
\end{tabular}
\caption{Percent HS and runtime comparison}
\label{tab:bulk_hs_runtime}
\end{table}

We summarize the result in Table \ref{tab:bulk_hs_runtime} and visualize the structure in Figure \ref{fig:bulkm}. A careful inspection of the MATLAB implementation released by Garner et al. reveals that the computational cost stems from a few implementation choices and the optimization framework itself. First, the code solves every iteration for all single cells and their compound‑pair combinations; this multiplies the number of finite‑element homogenizations by roughly an order of magnitude compared with a single-scale, single‑cell workflow. Second, the element‑wise density filter matrices are rebuilt inside the main loop, even though they depend only on mesh dimensions and filter radius. Third, each finite‑element system is factored afresh with MATLAB’s default sparse LU decomposition with further optimization potential. Finally, several inner loops are written in scalar MATLAB syntax, which limits vectorization and prevents automatic multithreading.

Taken together, these aspects explain why the reference implementation required 39 min to reach an average 77.4\% of the Hashin–Shtrikman upper bound for the bulk modulus optimization test, whereas our Python/TensorFlow implementation solves for a slightly extended FE problem per cell without significantly increasing the FE simulation to establish compatibility. Furthermore, the linear algebra solver and optimization were also implemented in completely different packages and environments. Thus, we completed an equivalent optimization in 74s and attained 82.5\% of the theoretical bound. However, the specific runtime performance increase requires further testing and potentially recreating Garner et al.'s code in the same environment as ours. 

%The disparity, therefore, reflects differences in environment, low‑level linear‑algebra handling, and efficiency in enforcing cell compatibility. 

\section{LIMITATIONS AND FUTURE WORK}

This study extends neural representation to multiscale and integrates it into heterogeneous metamaterial design. While promising, several challenges remain in the computational resources required for full-scale solving of multiscale optimization. Additionally, to maintain simplicity in implementation and sensitivity analysis, we deliberately chose a one-layer neural architecture. This architecture has demonstrated sufficient flexibility and offers a good trade-off between performance and computational efficiency for the problems studied in this paper. Further investigation of deeper networks and alternative architectures (e.g., Kolmogorov-Arnold Networks \cite{liu2025KAN}) tailored to specific applications represents a promising direction for future research.

\section{CONCLUSIONS}
In this work, we introduce a neural network-based approach for concurrent metamaterial design, providing a continuous and compact representation of two-scale structures. By using coordinate-based neural networks, we enable smooth transitions between unit cells while maintaining structural compatibility. We demonstrate the capability with popular metamaterial design problems, including displacement matching, NPR material design, mechanical cloaking, and compliance minimization. For problems involving target displacements, we observe good convergence with small errors. When designing with a base cell type, the neural network further allows a smoother transition between cell types compared to data-driven methods. Our method simplifies multiscale optimization by rephrasing the design problem as network weight optimization, thereby eliminating the need for discrete unit cell assignments. Additionally, the ability to query the trained model at any resolution allows for flexible refinement and scalability. The proposed framework addresses key challenges in metamaterial design, including compatibility enforcement and computational efficiency, while supporting the generation of seamless, high-resolution structures. 

\bibliographystyle{asmems4}
\bibliography{References}
\end{document}